\documentclass[10pt,twocolumn,letterpaper]{article}
\usepackage[hmargin=0.875in,vmargin=1in,columnsep=0.25in]{geometry}
\usepackage{times}
\usepackage{microtype}
\usepackage{graphicx}
\graphicspath{{./}}
\usepackage{booktabs}
\usepackage{xcolor}
\usepackage{xurl}
\usepackage[round,authoryear]{natbib}
\let\cite\citep

\usepackage[font=small,labelfont=bf,labelsep=period]{caption}
\usepackage[utf8]{inputenc}
\usepackage[T1]{fontenc}
\usepackage{amsmath,amssymb,amsthm}
\usepackage{array,longtable}
\usepackage{enumitem}
\usepackage{tikz}
\usetikzlibrary{arrows.meta,fit,positioning,shapes.geometric,calc}
\usepackage{hyperref}
\newtheorem{property}{Property}
\newtheorem{proposition}{Proposition}
\newcommand{\system}{FedReady}
\newcommand{\cohort}{\widehat{\mathcal{S}}}
\hypersetup{
  pdftitle={FedReady: Bounded Agents for Federation Formation},
  pdfauthor={Ziyue Xu, Holger R. Roth, Chester Chen, and Peter Cnudde},
  pdfsubject={arXiv preprint},
  pdfkeywords={federated learning, agentic systems, data preparation, verification},
  colorlinks=true,
  allcolors=black,
  hypertexnames=false
}
\title{FedReady: Bounded Agents for Federation Formation}
\author{Ziyue Xu \quad Holger R. Roth \quad Chester Chen \quad Peter Cnudde\\[0.4em]
NVIDIA\\[0.2em]
\small\texttt{ziyuex@nvidia.com}}
\date{\small Preprint}
\begin{document}
\twocolumn[{
\begin{minipage}{\textwidth}
\csname @twocolumnfalse\endcsname
\maketitle
\begin{abstract}
Federated learning (FL) runtimes standardize aggregation and deployment, but
each new study still requires \emph{federation formation}: selecting
task-eligible sites and preparing their local data and task code for federated
training or analysis. This preparation recurs when the target changes, while
raw records and site-private mappings are excluded from cross-site FL
messages. We present FedReady, a system that performs federation formation
with bounded LLM agents. Agents propose shared
contracts, local adapters, and task code; the runtime owns workflow state and
accepts artifacts only after checks tied to their digests. Candidate site
discovery prioritizes recall; local eligibility verification checks task
support before site admission. Once participating sites and artifacts are fixed, conventional
federated training or analysis proceeds without readiness agents.
Across 39 simulated retinal sites and four tasks, discovery reaches recall
1.00 at precision 0.65; local eligibility verification removes all 17 false positives
without losing an eligible site--task pair. The participating sites supply
45,575 task-record instances and complete four 100-round jobs, including one with a
runtime-generated contract. Across ten patient cohorts, preparation supports
seven patient populations and holds endpoints at three sources as unsupported.
Comparison with manually prepared reference populations and analysis (the manual reference) gives a
maximum survival-curve difference of 0.001133. We also analyze 19 development
failures, targeted classification ablations, and standalone and heterogeneous
agent configurations. These results establish the feasibility of task-specific
data preparation before ordinary FL training or analysis.
\end{abstract}
\end{minipage}
\vspace{1.5em}
}]

\section{Introduction}
\label{sec:intro}
Federated learning (FL) lets institutions train a shared model without
centralizing their records~\cite{mcmahan2017communication}. Production
runtimes handle aggregation, deployment, and fault
tolerance~\cite{bonawitz2019towards,roth2022nvflare}, but they start from a
federation that already exists: sites that can each emit records under one
shared interface and task code that consumes them. We define
\emph{federation formation} as selecting task-eligible sites and preparing
their local data and task code for federated training or analysis. Even within
an established FL alliance holding \emph{structured data}, preparing
data for a \emph{particular FL task} remains separate work:
defining a shared data contract, writing conversion code
against each site's native annotations or clinical fields, and producing task
code that runs on the result. This stage recurs \emph{per task}, not per
deployment: disc segmentation and disc detection over the same images need
different contracts, adapters, and trainers. Even across a 52-database network
that already shares a standardized schema, operators still report the need
for feasibility feedback and coordinated code
deployment~\cite{burkard2025federated}.

Automating this stage requires more than code generation.
The evidence that decides whether a site qualifies sits behind that site's
disclosure boundary: a local codebook that says whether a numeric code means
glaucoma, an annotation format that may or may not contain the requested
structure, a duration field whose endpoint is defined only in a source
publication. Under the configured disclosure policy, the FL server receives
approved summaries rather than raw records and site-private mappings;
public documentation can still inform interpretation. A common schema does
not supply the missing meaning.

LLM agents can now carry out much of this work~\cite{li2026helmsman,saha2026fedagentbench},
but handing it to them raises a placement question that model quality alone
does not answer. A server agent cannot verify a predicate over evidence it cannot
see. A site agent that emits plausible adapter code has not shown that the
code produces correct records. A reporting agent that finds an error in a log
has not shown that the job failed. Each is an ownership failure: a component
decides something whose evidence or authority belongs elsewhere.

\begin{figure*}[t]
\centering
\input{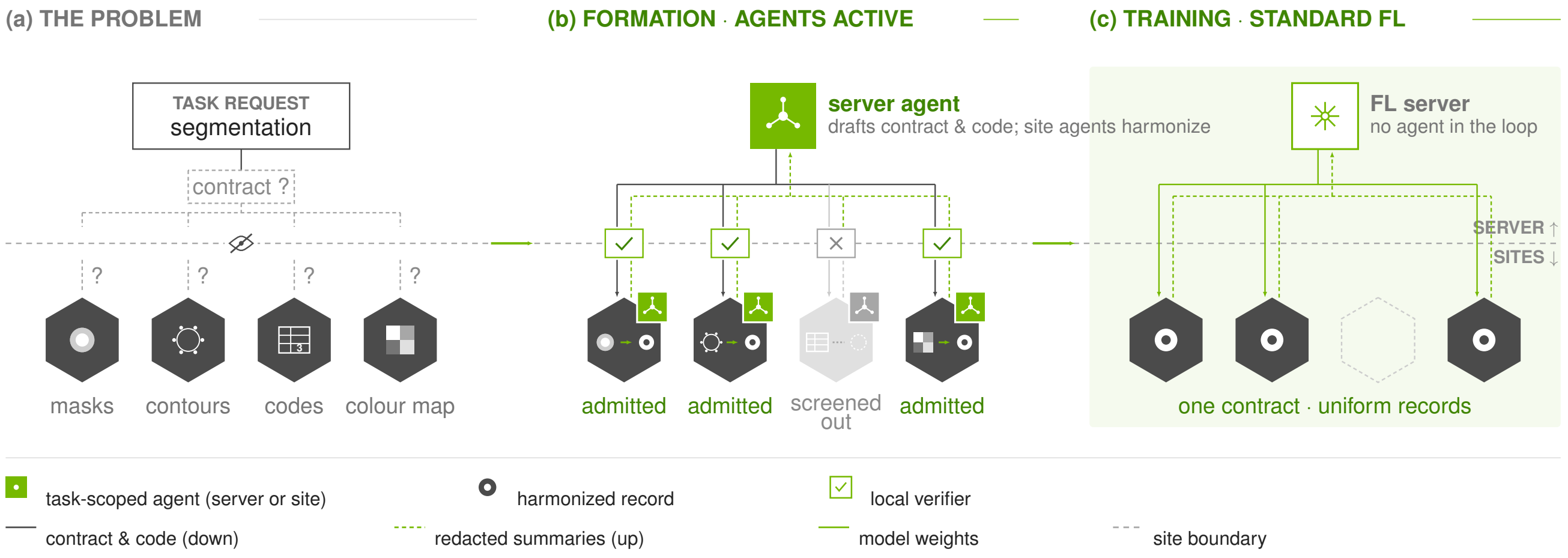}
\caption{Federation formation, illustrated for imaging. A new task encounters
incompatible private representations (a). Bounded agents propose a shared
data contract, local adapters, and task code; local verifiers check prepared
data before site admission (b). Ordinary FL then runs on fixed inputs,
without readiness agents (c). The hosted
coding-provider channel is not shown (Section~\ref{sec:disclosure}).}
\label{fig:overview}
\end{figure*}

\system{} performs federation formation
with agents bounded by \emph{view} (evidence they may observe), \emph{output}
(artifacts they may propose), and \emph{authority} (they cannot accept
artifacts or admit sites). Server agents draft the contract and task code;
site agents write adapters. The runtime owns transport, per-site state, and
artifact acceptance, requiring checks tied to each artifact's digest. With
participating sites and artifacts fixed, readiness agents stop and ordinary
FL training or analysis begins (Figure~\ref{fig:overview}).

The central design rule is to \emph{place each decision at the boundary that
owns its evidence}. Within a fixed formation episode, terminal exclusion loses
eligible sites; retained ineligible sites remain removable by later checks.
\system{} assigns high-recall
candidate site discovery to the server and local eligibility verification
to site verifiers that execute generated adapters against locally
supported semantics. On 39 simulated retinal sites and four tasks, this
pipeline screens out 17 ineligible candidates against the evaluator's
ledger, without sending raw evidence to the FL server. The hosted coding
provider is a separate trust boundary (Section~\ref{sec:disclosure}).

Our contributions are:
\begin{itemize}
\item \textbf{Abstraction.} We identify federation formation as a recurring,
per-task stage between private data and an FL runtime, with an audit graph
and four separately checkable validity criteria
(Section~\ref{sec:problem}).
\item \textbf{System.} We build \system{} on a production FL runtime, NVIDIA FLARE~\cite{roth2022nvflare}, around
runtime-owned artifact acceptance and evidence-based decision placement,
with request-bound messages, per-site states that cannot skip required checks,
task-code execution tests, and training from fixed verified inputs. An extension
shares verified procedures among heterogeneous site agents
(Section~\ref{sec:protocol}).
\item \textbf{Evaluation.} Across 156 imaging site-task decisions, local eligibility
verification removes all 17 discovery false positives without recall loss;
four 100-round jobs complete, one with a generated contract. Seven of ten
patient cohorts support analysis, with maximum survival-curve difference
0.001133 from the manual reference (Section~\ref{sec:evaluation}).
\item \textbf{Lessons.} An audit of 19 development failures classifies 15
primarily as ownership or state-integrity faults, yielding design lessons for
agentic ML pipelines (Sections~\ref{sec:failure-summary}
and~\ref{sec:discussion}).
\end{itemize}
Our sites are diverse public datasets simulated as private sites on one host. The
evaluation therefore tests logical boundaries, site-selection correctness, and
executability; it does not measure labor savings, formal privacy, or clinical
validity (Section~\ref{sec:discussion}). Field mappings, provenance policies,
and the incident index are in the appendix.

\section{Related Work}
\label{sec:related}
\begin{table*}[t]
\caption{Positioning relative to related agentic systems: primary objectives
and agent-produced outputs. \system{} connects task-specific site selection
and data preparation to runtime-controlled acceptance before training or analysis.}
\label{tab:related-systems}
\centering
\small
\setlength{\tabcolsep}{4pt}
\begin{tabular}{@{}>{\raggedright\arraybackslash}p{0.20\textwidth}>{\raggedright\arraybackslash}p{0.31\textwidth}>{\raggedright\arraybackslash}p{0.45\textwidth}@{}}
\toprule
Work & Primary objective & Main mechanism or output \\
\midrule
FedAgentBench~\cite{saha2026fedagentbench} & Automate and benchmark medical FL workflows & Server--client agents coordinate client selection, preprocessing, label harmonization, and training. \\
\addlinespace[2pt]
Helmsman~\cite{li2026helmsman} & Synthesize FL systems from user specifications & Agents generate plans and modular FL code, then evaluate and refine implementations in simulation. \\
\addlinespace[2pt]
Auto-FL-Research~\cite{roth2026autofl} & Search training algorithms within an existing federation & Coding agents propose and evaluate changes to FL training recipes. \\
\addlinespace[2pt]
FederatedSkill~\cite{yang2026federatedskill} & Improve agents through collaborative skill evolution & Agents share semantic skill patches to update personalized skill libraries. \\
\addlinespace[2pt]
This work (\system{}) & Select eligible sites and prepare task-compatible inputs & Agents propose contracts, local adapters, and task code; the runtime controls acceptance and site admission using local evidence and execution checks. \\
\bottomrule
\end{tabular}
\end{table*}

\paragraph{FL systems and data validation.}
FL runtimes standardize training and deployment
interfaces~\cite{bonawitz2019towards,roth2022nvflare}; benchmarks such as
FedScale supply workloads with shared task interfaces~\cite{lai2022fedscale}.
Data validation checks conformance~\cite{breck2019data}, yet federated
analyses still require coordination on standardized
data~\cite{burkard2025federated}. \system{} constructs task interfaces from
private sites; schema checks cannot establish codebook or endpoint meaning.

\paragraph{Agentic FL.}
FedAgentBench is closest because it also addresses client selection and data
harmonization~\cite{saha2026fedagentbench}; we reuse its retinal population
list. Our emphasis is on checking generated artifacts and proposed site
admissions against local evidence, runtime-controlled acceptance, and fixing
the accepted configuration before downstream execution
(Table~\ref{tab:related-systems}). Automation alone is not our novelty claim.
Agent registries describe \emph{published}
capabilities~\cite{raskar2025nanda,singh2025registries}, not contract-specific readiness.

\paragraph{Shared procedural knowledge.}
FederatedSkill shares semantic patches for personalized agent
evolution~\cite{yang2026federatedskill}; knowledge sharing alone is not novel.
Our extension targets local data readiness: contrastive lessons retain
disambiguating observations and receiver checks, with unchanged local
verifiers. Separating these effects remains future work.

\paragraph{Executable feedback, failures, and agent security.}
Coding-agent interfaces and test generation emphasize execution over
plausible text~\cite{yang2024sweagent,ahmed2025otter}. Failure taxonomies
classify multi-agent failures~\cite{cemri2025mast}; our audit tracks
ownership-boundary violations. Security and leakage benchmarks motivate
boundary restriction and whole-trace audit~\cite{zhang2025asb,elyagoubi2026agentleak}.
Our formation interfaces are neither a security benchmark nor a formal privacy mechanism.

\section{Federation Formation}
\label{sec:problem}
Let $\mathcal{S}=\{S_1,\ldots,S_n\}$ be candidate sites. Site $S_i$ owns
private state $x_i=(D_i,R_i,P_i,E_i)$: records, their representation and
semantics, disclosure policy, and execution environment. Given a task $q$,
formation produces
\begin{equation}
W=(C,\{A_i,M_i\}_{i\in\cohort},T,J,G),
\label{eq:workflow}
\end{equation}
where $C$ is the shared contract, $A_i$ a local adapter, $M_i$ its immutable
prepared-data manifest, $T$ shared task code, $J$ the exported job, $G$ an
audit graph, and $\cohort\subseteq\mathcal{S}$ the participating sites.
The server observes policy-approved projections $z_i=\rho_i(x_i)$, never
$x_i$ itself.

Two features make formation a coordination problem rather than a fixed
pipeline. First, \emph{capability is indexed by site and task}: whether a
site can supply a target depends on $q$. Published metadata may be
insufficient for a new target, so earlier site admission does not replace
contract-specific verification. Second, \emph{the decisive artifacts are executable}: a capability
claim means little until code at that site produces contract-valid records,
so discovery and code synthesis must be linked to execution-derived checks.

Success is the conjunction of four separately reported criteria, since none
implies the others: \emph{site-selection correctness} ($\cohort$ matches an
evaluator's eligible-site ledger), \emph{local validity} (each $M_i$ holds
contract-valid, provenance-bound records), \emph{package validity} ($T$
completes a simulation on \emph{synthetic test data}: contract-valid records
used for execution testing), and \emph{execution
validity} ($J$ completes the multi-client job and persists its outputs).
Correctly screening out a site whose evidence does not support the target is a
valid outcome, distinct from implementation failure. Our gates establish
conformance to implemented predicates, not an independent clinical oracle.

\paragraph{Trust model and scope.}
We assume honest-but-fallible agents, malformed generated artifacts,
accidental over-disclosure, ambiguous source semantics, and backend or
transport failures. The FL runtime is a centralized control plane; evidence
stays decentralized at the sites. We do not defend against a malicious host,
provide OS-grade confinement of generated code, or establish differential
privacy. In the evaluated configuration a hosted coding worker operates under a
site-scoped access policy, so keeping records out of cross-site messages does
not prove non-observation by its provider.

\subsection{Artifact Acceptance and Site Admission}
\label{sec:properties}
\emph{Artifact acceptance} approves generated contracts or code;
\emph{site admission} accepts verified sites into the federation; a \emph{state transition}
records workflow progress. These are separate runtime decisions.
Three invariants, each enforced by an interface rather than a prompt,
organize the design:
\emph{scope} (I1: a record-level observation cannot silently become a site-level
veto), \emph{proposal} (I2: agent output is never a verifier result), and
\emph{terminal state} (I3: the runtime's completion evidence, not a narrative or
log token, determines success).

\begin{property}[Acceptance tied to checked artifacts]
\label{prop:soundness}
Assuming an honest host and intact acceptance interfaces, every accepted
contract, adapter/manifest, and trainer carries a runtime-owned attestation
(verification record) bound to the digest (content hash) of its reviewed bytes.
\end{property}
The runtime restores verification fields from its own checks and accepts only
artifacts with the reviewed digest. Agent self-reports cannot replace these checks; visual
observations may inform them but cannot authorize acceptance. This argument
bounds \emph{who} accepts artifacts, not semantic correctness; it is not a
machine-checked theorem.

The second result concerns \emph{where} a decision is taken. For the eligible
set $\mathcal{E}$ and shrinking active site sets
$\mathcal{S}_0\supseteq\cdots\supseteq\mathcal{S}_K=\cohort$, define recall as
$|\mathcal{S}_k\cap\mathcal{E}|/|\mathcal{E}|$.
\begin{proposition}[Authority placement]
\label{prop:placement}
For fixed $\mathcal{E}$, terminal exclusion without re-entry means recall cannot increase along the
stage chain. An excluded eligible site is unavailable to later checks; a
retained ineligible site remains removable by a later evidence-sufficient
verifier.
\end{proposition}
This follows from set inclusion. A stage seeing only $\rho_i(x_i)$ should
defer predicates over omitted evidence to a later sound and complete verifier;
deferral alone guarantees nothing. Retries repair implementations, not
justified screening. Changed evidence requires a new formation episode;
live re-admission is unevaluated.
Section~\ref{sec:protocol} implements this split, and
Section~\ref{sec:cohorts} measures both of its terms.

\section{The FedReady System}
\label{sec:protocol}
\system{} implements Section~\ref{sec:properties} through candidate site discovery, contract
resolution, local adaptation, and task-code execution tests before ordinary FL.
Only the runtime authorizes state transitions.

\subsection{Architecture and Runtime Ownership}
Agents receive request-scoped work items, not controller handles or peer
addresses. They can propose artifacts but cannot invoke another site, mutate
the participating-site list, or approve their own artifacts
(Table~\ref{tab:roles}).

\begin{table*}[t]
\caption{Roles are assigned by evidence access and permitted state effects,
not by a particular backbone. Agents propose artifacts; the runtime accepts them after checks.}
\label{tab:roles}
\centering
\small
\begin{tabular}{@{}>{\raggedright\arraybackslash}p{0.16\textwidth}>{\raggedright\arraybackslash}p{0.35\textwidth}>{\raggedright\arraybackslash}p{0.43\textwidth}@{}}
\toprule
Role & Permitted view/output & Authority boundary \\
\midrule
FL runtime & Typed messages, contract, site list, digests; dispatch, checks, export, training & Owns transport, state, verification records, and terminal evidence; does not infer private label meaning \\
Server planner/coder & Task, redacted summaries, synthetic test data; inquiry, contract, task code & No raw site evidence, direct client calls, or artifact acceptance \\
Site adapter coder & Authorized local structure, documentation, contract; adapter or abstention & No sibling access, contract mutation, or visual-verdict ownership \\
Local visual witness & Bounded raw input and contract-owned rendering; structured observations & No cross-site images, code edits, or site admission \\
Local verifiers & Owning-site artifacts; structure, provenance, and execution checks & No invented semantics, silent repairs, or reference-code substitution \\
Disclosure filters & Message payloads; field checks and redaction & No artifact acceptance or site admission \\
\bottomrule
\end{tabular}
\end{table*}

\textbf{What the runtime must provide.}
NVIDIA FLARE~\cite{roth2022nvflare} owns transport, state, and execution.
Typed \texttt{Shareable} headers bind replies to requests; controllers and
executors own cross-site dispatch. The simulator tests exported jobs against
synthetic test data, and sample-weighted FedAvg runs imaging training.
These are logical workflow boundaries, not deployment-grade isolation.
Another runtime would need equivalent state ownership, request binding, and
deployment-faithful simulation.

\textbf{Backbones.}
In the primary imaging configuration, hosted Codex workers perform global planning,
task-code synthesis, and site-scoped adapter coding, and
Qwen3-VL-8B-Instruct supplies bounded visual judgments at a local endpoint.
Separate Luna/Codex and Pi/Nemotron studies extend the preparation evaluation;
Table~\ref{tab:experiment-configurations} records campaign-specific model,
harness, effort, and budget settings.

\subsection{Typed State and Disclosure Boundaries}
Within an attempt, each site advances through gated states:
\emph{registered}, \emph{profiled}, \emph{plausible}, \emph{contract-bound},
\emph{adapter-attested}, \emph{materialized}, \emph{verified}, and
\emph{train-ready}. Screened-out sites, held records or endpoints, and
implementation failures have distinct typed dispositions. Generated-contract tasks add a
\emph{materializer-ready} gate. A repair creates a new, bounded attempt; it
cannot overwrite an attestation or skip a gate.

Every reply is bound to session, correlation, schema, task, site, round, and
action, including the payload's client identity, so a valid JSON object from
the wrong request cannot trigger a state transition. The transcript links artifact
versions, checks, and dispositions into the audit graph $G$ instead of
treating free-form logs as the state record.

\label{sec:disclosure}
Disclosure has two channels. Cross-site FL messages exclude records,
codebooks, source paths, adapter source, and manifests, using typed status,
counts, reason codes, booleans, digests, and authorized summaries instead.
Optional semantic redaction preserves required fields and is runtime-audited.
Separately, the hosted coding provider receives prompts and tool-visible
content: local file reads can expose documentation, paths, record excerpts,
code, and diagnostics. Site-scoped tools do not imply local model inference;
a local VLM does not protect this coding channel. We have not
audited complete provider-bound traffic. These are coarse flow
controls~\cite{sabelfeld2003language}, not a privacy guarantee.

\begin{figure*}[t]
\centering
\begin{minipage}{0.94\textwidth}
\centering
{\footnotesize
\makebox[0.25\linewidth]{\textbf{(a) Binary-mask overlay}}%
\makebox[0.25\linewidth]{\textbf{(b) Disc/cup overlay}}%
\makebox[0.25\linewidth]{\textbf{(c) Edge-index example}}%
\makebox[0.25\linewidth]{\textbf{(d) Binary disc mask}}\par}
\includegraphics[width=\linewidth]{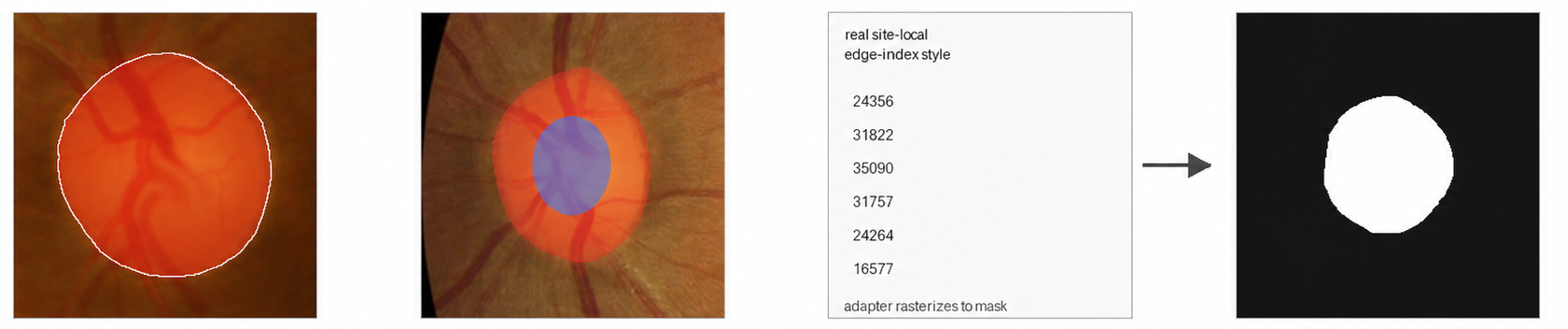}
\end{minipage}
\caption{Different annotation representations, not successive views of one
case: (a) a binary disc-mask overlay; (b) a multi-valued mask overlay
(orange disc, purple cup); (c) an edge-index illustration derived from a real mask
boundary; (d) the binary disc-mask format required by the data contract.
The arrow illustrates rasterization of boundary indices. Adapters must resolve
encoding and label meaning; Table~\ref{tab:tabular-challenges} shows the tabular counterpart.}
\label{fig:local-translation}
\end{figure*}

\subsection{Candidate Site Discovery and Local Eligibility Verification}
\paragraph{Candidate site discovery.}
To find potentially suitable sites, the server broadcasts a bounded inquiry
for $q$. Sites profile authorized structure and may use agents to interpret
local documentation or codebooks. Redacted site profiles support
\emph{plausibility}, not verified eligibility. Discovery prioritizes recall,
retaining candidates for local eligibility verification
(Proposition~\ref{prop:placement}), and can stop early if no site reports
plausible data.

\paragraph{Contract resolution.}
The server resolves a versioned \emph{data contract} from $q$ and redacted site
profiles. The contract specifies required structure, meaning, and constraints;
its \emph{schema} defines the structural part. The \emph{interface} is how task
code consumes the prepared records under that contract.
Built-in contracts cover binary segmentation and image-level classification.
For a new task family, an agent proposes a contract, and a separate
request produces a materializer that converts adapter outputs to its schema.
Required fields, types, invariants, and synthetic test data construction are checked
before dispatch. The contract owns the invariants, so generated code cannot
invent exclusion rules, and adapter execution binds to the selected contract
and materializer digests.

\paragraph{Local eligibility verification.}
A site agent checks task support against authorized inputs and proposes an adapter
and manifest. The runtime snapshots pre-existing source identities, compiles
the adapter, and runs \emph{sample adapter checks} on at most eight real local
records: declared semantics, pairing, types, conversions, and exact
source/manifest binding. Newly generated annotations cannot masquerade as
original label evidence, and live modes never fall back silently to a
reference adapter.

The server receives typed extraction outcomes and aggregate verification
summaries. A \emph{site is screened out} when its evidence cannot support the
task. A \emph{record or endpoint is held} when source evidence or contract
rules do not support its inclusion in analysis. Missing target values have a
separate status. These outcomes differ from \emph{implementation failure}:
required code or its execution fails, which does not establish data ineligibility.
Correct screening or holds count as completed preparation, not workflow failure.

\textbf{Medical imaging data.}
For image-bearing tasks requiring visual evidence, a local visual witness observes a bounded raw input and
a contract-owned rendering, conditioned on a digest-bound task example; its
verdict is a structured observation, not permission to edit code or admit a
site. Full materialization then checks required fields, counts, label
domains, splits, readability, and provenance. A representative check does not
establish dataset-wide visual or clinical correctness. Broad quality grading
and autonomous relabeling are outside this paper.

\textbf{Clinical tabular data.}
For tables, the contract also states the analysis unit, endpoint, event
coding, duration units, and population-selection rules. Agents inspect
documentation and propose site mappings and measurable assertions; local
programs then apply those policies to all records without per-record LLM
review. Fixed coordinator checks cover identifiers, target domains, hashes,
and contracted predicates over execution-derived measurements, and an
aggregate reviewer can request local corrections or a new contract. The same
distinction between proposal and artifact acceptance applies: assertions test conformance to an
issued interpretation but cannot independently establish its clinical
meaning.

\subsection{Task-Code Execution Tests and Training Handoff}
After local preparation, the server coder receives the contract, approved
aggregates, and synthetic test data, never real site records,
and proposes a contained task package. For imaging training, schema, path,
containment, and compilation checks precede \emph{task-code execution tests}:
a runtime-owned one-client simulation on synthetic test data checks the
exported command line, imports, data loader, model,
optimizer, update exchange, metrics, and persistence. Bounded failure
feedback returns to the same coding role for repair. Passing these tests
is neither operator deployment approval nor clinical validation.

The runtime then fixes $C$, $\cohort$, $\{M_i\}$, and the task-code digest,
and ends readiness-agent participation. Each admitted client re-resolves its
prepared-data root and checks manifest, schema, splits, and digests before
running shared task code; the exported job contains no prepared examples or
per-record manifests. Imaging rounds exchange model updates and approved aggregate
metrics under sample-weighted FedAvg~\cite{mcmahan2017communication}. The
tabular application instead aggregates endpoint-specific event/censor
histograms with a numerical check, without training a predictive model. In
both workloads there are no readiness agents or shared-guidance updates downstream.
The present simulation skips human approval; deployment would require
operator approval and stronger code confinement.

\subsection{Extension: Heterogeneous Site Agents}
\label{sec:heterogeneous-design}
The base \textbf{FedReady} uses one site-agent configuration. Agents
introduce another form of FL heterogeneity, \emph{agent heterogeneity}, in
which each site can use a different agent backbone. We therefore
investigate \textbf{FedReady-Hetero}, which permits different site-agent
configurations. The extension
keeps the same contracts and runtime checks and adds shared guidance.
Initially successful stronger agents share procedures intended to
make weaker agents' preparation more consistent.

\textbf{Local contrastive-guidance generation.}
After verified preparation, a designated contributing site abstracts its
procedure into shared guidance with site-specific details removed. A \emph{contrastive lesson} records a
plausible incorrect interpretation, the local observation that rules it out,
and a check the receiver must perform; for example, an output mask must not
be mistaken for a pre-existing annotation source. Lessons derive from verified
success or an actually resolved failure and cannot invent a failure history.

\textbf{Global distillation.}
The server distills validated lessons into common steps and conditional
procedures gated on receiver-local evidence, removing donor paths, examples,
adapters, numeric codebooks, and unresolved conflicts. Lessons explain
\emph{how to investigate}, not another site's label meanings. Provenance
checks do not prove correctness or absence of leakage. Site agents first attempt
independently; unresolved implementations may retry with shared guidance under
unchanged local verifiers, and verified successes may contribute lessons.
Completed sites do not rerun; justified screening is not reversed. Retries stop
when preparation completes, no new shared guidance is produced, or the
attempt limit is reached. The aim is
repeatable success within existing capability, not guaranteed rescue
(Section~\ref{sec:guidance-stability}).

\section{Evaluation}
\label{sec:evaluation}
We ask five questions across two modalities:
\textbf{RQ1} (site admission): do candidate site discovery and local eligibility
verification select the correct sites under the disclosure policy?
\textbf{RQ2} (new task interface): can the fixed protocol build and execute
a new task interface?
\textbf{RQ3} (training and analysis): do accepted packages execute, and how
closely does tabular analysis match the manual reference?
\textbf{RQ4} (agent configurations): how does readiness vary across agents
and harnesses?
\textbf{RQ5} (failures): which boundaries do development failures violate?
Imaging and tabular studies use distinct coordinators, not one combined run.

\subsection{Workloads and Methodology}
\begin{table*}[t]
\caption{Representative native-column challenges in the clinical workload.
GEO names are characteristic keys; TCGA prefixes denote nested objects.
Additional evidence determines which fields and records support the task,
or whether targets must be held.}
\label{tab:tabular-challenges}
\centering
\small
\setlength{\tabcolsep}{4pt}
\begin{tabular}{@{}>{\raggedright\arraybackslash}p{0.12\textwidth}>{\raggedright\arraybackslash}p{0.28\textwidth}>{\raggedright\arraybackslash}p{0.32\textwidth}>{\raggedright\arraybackslash}p{0.22\textwidth}@{}}
\toprule
Source & Native columns & Task selection and interpretation & Additional evidence \\
\midrule
GSE14333 & \nolinkurl{DFS_Time}, \nolinkurl{DFS_Cens}, \nolinkurl{DukesStage} &
DFS for supported Dukes B/C; months/12; event $=1-\texttt{DFS\_Cens}$. Completeness alone does not confer eligibility. &
Published methods for this patient cohort~\cite{jorissen2009metastasis}, recurrence counts, duration distributions. \\
\addlinespace
Ethiopia & \nolinkurl{TIME}, \nolinkurl{VSTATUS}, \nolinkurl{statusfn}; \nolinkurl{Date_Dx}, \nolinkurl{Date_death} &
CRC-specific survival, not all-cause OS; months/12; 1 means CRC death. Hold contradictory chronology. &
Endpoint definition~\cite{teka2021colorectal}, duplicate status, published person-months/life table. \\
\addlinespace
TCGA COAD/READ & \texttt{demographic}: \nolinkurl{vital_status}, \nolinkurl{days_to_death}; \texttt{diagnoses}: \nolinkurl{days_to_last_follow_up}; \texttt{follow\_ups}: \nolinkurl{days_to_follow_up} &
Case-level OS: death days for dead cases; latest applicable nonnegative follow-up for alive cases; days/365.25. Follow-up cannot replace missing death time. &
GDC dictionary/release guidance~\cite{gdcClinicalDocumentation}; clinical outcome methods~\cite{liu2018clinical}. \\
\addlinespace
GSE35896 & Mirror \nolinkurl{ID_REF}, \texttt{DFS (in months)}, \texttt{DFS event}, \nolinkurl{Gender}; GEO \nolinkurl{^SAMPLE} &
All 62 accessions join, but targets remain held: linkage does not authenticate secondary clinical fields. &
Primary-source comparison; 32 joined-record sex conflicts and unresolved mirror provenance. \\
\bottomrule
\end{tabular}
\end{table*}

\paragraph{Retinal imaging data.}
We register 39 public retinal packages, following the population listed by
FedAgentBench~\cite{saha2026fedagentbench}, as simulated private sites. They
differ in layout, annotation representation, codebooks, and task eligibility.
Four tasks (disc segmentation, cup segmentation, glaucoma classification, and
disc detection) are each issued to all 39 sites, giving 156 site-task
decisions. Figure~\ref{fig:local-translation} contrasts binary masks,
multi-valued masks, and boundary indices: adapters must interpret the native
labels and produce the binary disc-mask interface required by the task.

\paragraph{Clinical tabular data.}
\label{sec:tabular}
Ten public colorectal-cancer patient cohorts (seven GEO series, TCGA colon and
rectal cohorts, and an Ethiopian cohort) contain 2,189 source records,
including controls. The authors separately prepared reference populations
and analysis (the manual reference). Both methods use the same checksum-bound
pool of 98 clinical-metadata and
documentation files; expression matrices, manual adapters, prepared reference
tables, and reference outputs are unavailable to live agents. The task is to
construct supported populations for endpoint-specific survival analysis, not
to pool different endpoints under one schema. As
Table~\ref{tab:tabular-challenges} shows, a value can have the right type but
the wrong endpoint, event polarity, time origin, analysis unit, or
population. Identity, applicability, and eligibility must precede
target-completeness filtering, and a supported target can coexist with an
unsupported covariate. A dedicated tabular coordinator reuses the
architecture and agent bridge of Section~\ref{sec:protocol}, not the
unchanged imaging controller.

\paragraph{Task specification and automation boundary.}
Table~\ref{tab:automation-boundary} separates supplied inputs from generated
artifacts. Fixed runtime checks cover schemas, provenance, artifact binding,
and executable conformance. For tables, the coordinator also evaluates
agent-generated measurements and assertions over executed outputs. These can
repeat an adapter's mistaken assumption: passing is not independent semantic
verification. Agents formulate policies; local programs process records.

\begin{table}[t]
\caption{Automation boundary across workloads. Task requests, source inputs,
and shared runtime checks are supplied in every case.}
\label{tab:automation-boundary}
\centering
\small
\setlength{\tabcolsep}{3pt}
\begin{tabular}{@{}>{\raggedright\arraybackslash}p{0.22\columnwidth}>{\raggedright\arraybackslash}p{0.32\columnwidth}>{\raggedright\arraybackslash}p{0.40\columnwidth}@{}}
\toprule
Workload & Supplied specification & Agent-produced artifacts \\
\midrule
Segmentation; classification & Built-in data contract & Site mappings, adapters, task code \\
\addlinespace
Detection & Generic field/storage checks & Contract, materializer, local conversions, trainer \\
\addlinespace
Clinical tables & Output fields and screening categories & Endpoint/selection rules, adapters, measurable assertions, analysis code \\
\bottomrule
\end{tabular}
\end{table}

\paragraph{Reference evidence and metrics.}
Both studies run public sources as private-site stand-ins on one host. The
tabular clients are source patient cohorts, not ten verified independent hospitals;
cross-source patient overlap is unresolved. For imaging, an evaluator-owned ledger, never exposed to agents or used as an
admission allow-list, defines eligibility from locally available evidence:
spatial tasks need provable target supervision, and classification needs
explicit local diagnosis meaning rather than structural proxies or
undocumented numeric codes. We report discovery precision and recall, final
site admissions, and completion of 100-round jobs. Prediction metrics are
unweighted means of available client validation scalars, not pooled
patient-level scores. For tables, we report supported versus held endpoints by source,
endpoint-specific record and event counts, monthly survival-curve agreement
with the manual reference, and a separately implemented arithmetic check on
the agent's accepted inputs. Neither study measures a matched speedup or predictive
superiority over a baseline.

\subsection{RQ1: Decision Placement and Site Admission}
\label{sec:cohorts}
\paragraph{Imaging admission.}
Candidate site discovery retained 49 site-task pairs from redacted site profiles
against 32 eligible pairs: recall 1.00 at precision 0.65. Local eligibility
verification removed all 17 false positives without recall loss: precision
1.00 at recall 1.00 (Table~\ref{tab:formation}). Admitted sites supply
45,575 materialized records across 32 task memberships, not 32 distinct
organizations. The server received
typed screening outcomes and approved summaries, not raw rejection evidence.
The targeted classification ablations below test the contribution of
individual preparation mechanisms.

\begin{table}[t]
\caption{Federation formation across four tasks, each issued to 39 sites.
Candidate counts include discovery precision in parentheses. Local checks
screen out all 17 ineligible candidates. Records reused across tasks count again.}
\label{tab:formation}
\centering
\small
\setlength{\tabcolsep}{2pt}
\begin{tabular}{@{}>{\raggedright\arraybackslash}p{0.28\columnwidth}crrr@{}}
\toprule
Task & \shortstack{Candidates\\(precision)} & \shortstack{Screened\\out} & Admitted & Records \\
\midrule
Disc segmentation & 12 (0.67) & 4 & 8 & 6,482 \\
Cup segmentation & 8 (0.75) & 2 & 6 & 6,062 \\
Glaucoma classification & 19 (0.53) & 9 & 10 & 26,549 \\
Disc detection & 10 (0.80) & 2 & 8 & 6,482 \\
\midrule
Total & 49 (0.65) & 17 & 32 & 45,575 \\
\bottomrule
\end{tabular}
\end{table}

Discovery precision is lowest for glaucoma (0.53), whose labels can require a
local codebook omitted from central profiles. One source provided documented
spatial targets but was screened out for classification: its numeric diagnosis
columns lacked a codebook. Four disc candidates lacked usable
target labels or pairings. Discovery precision 0.65 depends on this population
and metadata policy; richer approved metadata could change it.

\textbf{Targeted classification ablations.}
The two ablations use a matched control, not the historical training run
(Table~\ref{tab:classification-ablation}). The semantic-check arm retains
agents, inputs, instructions, local budgets, reasoning, and abstention across
19 candidates. It yields no false admissions, but lower token consumption
coincides with an eligible-site exclusion. Discovery is intended to avoid
contract-building and adapter effort when no client supports the task.

\begin{table}[t]
\caption{Glaucoma preparation: ten eligible sites, one run per configuration.
Tokens normalized to the matched full-workflow control.}
\label{tab:classification-ablation}
\centering
\small
\setlength{\tabcolsep}{2pt}
\begin{tabular}{@{}>{\raggedright\arraybackslash}p{0.44\columnwidth}rrr@{}}
\toprule
Configuration & \shortstack{Incorrect\\admissions $\downarrow$} & \shortstack{Eligible\\excluded $\downarrow$} & \shortstack{Relative\\tokens $\downarrow$} \\
\midrule
Full workflow & 0 & 0 & 1.000 \\
Without semantic-evidence and class-coverage checks & 0 & 1\textsuperscript{*} & 0.894 \\
Without candidate site discovery & 0 & 0 & 1.099 \\
\bottomrule
\end{tabular}
\par\smallskip
\begin{minipage}{\columnwidth}
\footnotesize
End-to-end tokens (row order): 10,467,482; 9,359,084; 11,500,395.
\textsuperscript{*}Provenance-related exclusion; causal link to removed checks unresolved.
\end{minipage}
\end{table}

\paragraph{Clinical population construction.}
All ten tabular workflows completed; seven supplied supported populations.
Endpoints at GSE33113, GSE35896, and GSE37892 remained held for endpoint,
identity, or provenance uncertainty. At GSE37892, executable date subtraction
cannot establish event-free follow-up from missing metastasis dates;
retaining only known metastases would select on the outcome. Holding an
unsupported endpoint completes preparation without including it in analysis.
One corrective cycle addressed GSE37892 and TCGA\_COAD; a TCGA\_READ adapter
failure was recovered. These outcomes distinguish unsupported semantics from
repairable implementation failure: schema validity or a successful join
alone cannot justify site admission or endpoint inclusion.

\subsection{RQ2: Adapting to a New Task Interface}
Disc segmentation and detection share eight sites and 6,482 images, but
require different schemas, materializers, and adapters: native boundaries
must become boxes for detection. Prior site admission does not establish
new-interface conformance; source evidence and artifacts require rechecking.

Detection was also a generated-contract probe. The runtime contains generic
field and storage validation, but no checked-in detection contract,
materializer, renderer, adapter, or fallback trainer. From the request and
redacted site profiles, the server generated an image-box contract and materializer,
eight site agents converted their own contours or masks into locally verified
boxes, and the server coder produced a Faster R-CNN-style package that passed
contract-preserving simulation and completed 100 rounds. Three interfaces
(shared schema, local conversion, shared trainer) changed at run time while
authority, message binding, and gates stayed fixed. This is one task-family
probe, not evidence of arbitrary task generality.

\subsection{RQ3: Training and Analysis on Verified Inputs}
\paragraph{Imaging training.}
All four jobs completed 100 rounds of sample-weighted FedAvg
(Table~\ref{tab:downstream-execution}). Detection's round-96 communication
warning did not prevent completion or persisted terminal artifacts. Learning
curves remain in the supplement as execution diagnostics.

\begin{table*}[t]
\caption{End-to-end execution after federation formation. Agent-generated
local adapters and shared task code support four completed FL jobs,
including detection with an agent-generated data contract. Participating
sites and artifacts remain fixed, with no readiness-agent intervention.
Validation metrics are unweighted means across reporting clients.}
\label{tab:downstream-execution}
\centering
\small
\setlength{\tabcolsep}{4pt}
\begin{tabular}{@{}llrcl@{}}
\toprule
Task & \shortstack{Shared data\\contract} & \shortstack{Participating\\sites} & \shortstack{Rounds\\completed} & Final validation result \\
\midrule
Disc segmentation & Predefined & 8 & 100/100 & Dice: 0.9383 \\
Cup segmentation & Predefined & 6 & 100/100 & Dice: 0.7740 \\
Glaucoma classification & Predefined & 10 & 100/100 & Accuracy: 0.8827; AUC: 0.9150 \\
Disc detection & Agent-generated & 8 & 100/100 & GT recall@0.5: 0.9316 \\
\bottomrule
\end{tabular}
\end{table*}

\paragraph{Survival analysis and agreement.}
A generated FLARE application, accepted on its second attempt, computes
separate monthly-grid Kaplan-Meier curves~\cite{kaplan1958nonparametric} from
local event/censor histograms after prepared inputs are fixed. Five clients
contribute overall survival (OS); disease-free (DFS) and CRC-specific
survival each use one client, so those two endpoints test endpoint-preserving
execution rather than multi-client aggregation.

\begin{table}[t]
\caption{Survival results (records/events). Curve differences compare agent
and manual populations on the same monthly grid.}
\label{tab:tabular-results}
\centering
\small
\setlength{\tabcolsep}{3pt}
\begin{tabular}{@{}lrrrr@{}}
\toprule
Endpoint & Sites & Agent & Manual & Max.\ $|\Delta S|$ \\
\midrule
OS & 5 & 1,410/410 & 1,417/412 & 0.001133 \\
DFS & 1 & 185/48 & 185/48 & 0 \\
CRC-specific & 1 & 160/74 & 160/74 & 0 \\
\bottomrule
\end{tabular}
\end{table}

DFS and CRC-specific counts and curves match at all 80 and 49 monthly points
(Table~\ref{tab:tabular-results}). OS differs in at least one
risk/event/censor count at 42 of 153 points, with maximum survival difference
0.001133 (0.1133 percentage points). Archived attribution lists four stage-0
and four prior-treated records held only by the agent, and one primary-case
record included only by the agent: seven fewer patients and two fewer events
overall. These are selection disagreements, not adjudicated errors; the
patient-row-free archive prevents independent replay. A separate arithmetic
check reports zero curve error on accepted inputs, not validation of original
eligibility or endpoint meaning. The manual reference is not independent
clinical adjudication.

\subsection{RQ4: Agent Configurations and Readiness}
\label{sec:heterogeneous-results}
FedReady-Hetero uses a Sol server, 20 Sol-high and 19 Luna-low site agents;
high/low denote reasoning effort. Table~\ref{tab:heterogeneous} reports
readiness, not training completion; initial R2 means round 2, not one-shot success.

\begin{table}[t]
\caption{Heterogeneous readiness: Sol-high / Luna-low counts, not fractions;
39 sites per task. ``- / -'': initial readiness retained without revision.}
\label{tab:heterogeneous}
\centering
\small
\begin{tabular}{@{}lccc@{}}
\toprule
Task & Expected & Initial R2 & Latest \\
\midrule
Disc segmentation & 4 / 4 & 4 / 2 & 4 / 4 \\
Glaucoma classification & 5 / 5 & 5 / 2 & 5 / 4 \\
Cup segmentation & 3 / 3 & 3 / 2 & 3 / 3 \\
Disc detection & 4 / 4 & 4 / 4 & - / - \\
\bottomrule
\end{tabular}
\end{table}

Readiness rises from $26/32$ to $31/32$ task memberships: Luna-low from
$10/16$ to $15/16$, Sol-high unchanged at $16/16$. One glaucoma membership
remains unresolved; detection needs no revision.

All four Luna detection sites succeed initially, but intentional Sol/Luna
reassignment between detection and disc segmentation prevents paired inference
about standalone segmentation success. Causal guidance effects, stability,
and record correctness require matched repeated trials and complete attempt
and guidance-use records (Section~\ref{sec:guidance-stability}).

\paragraph{Standalone agent configurations.}
All-Luna-low/Codex without shared guidance and Nemotron 3.5 Super/HIGH
with Pi each miss five eligible task memberships
(Table~\ref{tab:standalone-agents}); their missed sites do not overlap within
any task. Nemotron's misses are
implementation failures; Luna's failure types are unspecified. These separate
campaigns do not isolate model, harness, or guidance effects.

\begin{table}[t]
\caption{Standalone preparation on 39 sites per task: eligible memberships
missed against the same reference. Both configurations have zero incorrect admissions.}
\label{tab:standalone-agents}
\centering
\small
\setlength{\tabcolsep}{3pt}
\begin{tabular}{@{}lrrr@{}}
\toprule
Task & Eligible & \shortstack{Luna\\missed} & \shortstack{Nemotron\\missed} \\
\midrule
Disc segmentation & 8 & 1 & 0 \\
Cup segmentation & 6 & 1 & 2 \\
Glaucoma classification & 10 & 2 & 1 \\
Disc detection & 8 & 1 & 2 \\
\midrule
Total & 32 & 5 & 5 \\
\bottomrule
\end{tabular}
\end{table}

\subsection{RQ5: What Failed in Development}
\label{sec:failure-summary}
Archive review identifies 19 author-coded incidents: five authority, five
scope, five state-loss, and four environment failures; 15/19 concern ownership
or state integrity. Eight have a validating retry or regression test (E1);
eleven have a timestamped trace and diagnosis (E2). Four quality-screening
incidents (WF-002, WF-011, WF-012, WF-013) are out of scope. Without a second
annotator or reliable run/attempt denominator, incident shares are not
failure rates. This audit covers the primary coding configuration, not a
matched multi-backend study.

WF-001 treated \texttt{seg} as supervision and omitted class-folder evidence,
silently excluding an eligible 4,217-record classification source from a
completed run (Proposition~\ref{prop:placement}); narrowing parser ownership
recovered its verified manifest. In WF-004, synthetic test data lost approved
detection-contract fields; seven misdirected trainer revisions failed, while
preserving the contract enabled task-code execution tests and training.
These cases identify concrete interface-level repairs; incident summaries
and evidence grades are in Appendix~\ref{sec:failures}.

Other incidents involved undeclared materializer invariants, provenance
confusion, and log-based completion errors. All four environment faults
required exported-code execution to detect.

\section{Discussion and Limitations}
\label{sec:discussion}
\paragraph{Lessons for agentic ML pipelines.}
Four lessons follow: \emph{place authority by evidence}; \emph{execute the
exported interface}; \emph{repair the owning component}; and \emph{audit
transitions}. Typed outcomes distinguish screened-out sites, held records
or endpoints, and implementation failures, including backend failures.

\paragraph{Guidance as stabilization, not capability rescue.}
\label{sec:guidance-stability}
Shared guidance may stabilize an agent capable of independent but unreliable
success, not rescue persistent incapability. Testing this hypothesis requires
repeated, equal-budget trials with and without shared guidance, establishing standalone
success and comparing verified success rates. Zero successes in finite controls
do not prove incapability; this checkpoint establishes neither stability
gains nor a necessary capability threshold.

\paragraph{Matching fields can represent different clinical outcomes.}
Duration/event fields may differ in endpoint, time origin, or population.
Contracts must state these meanings and unsupported cases; curve agreement
cannot replace source-to-output population checks.

\paragraph{Potential time and effort saving.}
Clinical-data transformation and loading took over six person-months initially,
versus one to two for a subsequent health system with partially transferable
code and knowledge~\cite{fitzhenry2015cdm}; different sources preclude a causal
reuse estimate.
Nursing flowsheet mapping logged 95.33 concept-mapping and 74.75 value-mapping
hours~\cite{austin2025flowsheet}. FLEX-LLM-CARE estimates a reduction from 360 to 149 expert-hours
(59\% reduction) from historical timelines and expert feedback, retaining
40 hours for semantic validation~\cite{marfoglia2025flex}.
These reports motivate a matched preparation study rather than serving as
per-adapter baselines for \system{}.

Across 52 databases in 19 countries, partners typically ran code steps within
one to two months of deployment; execution took hours to days, or weeks with
issues~\cite{burkard2025federated}. Coordination delays were not isolated.
In our experience, response waiting, time zones, meeting
scheduling, and clarification can exceed conversion time. Matched human,
coding-assisted, and \system{} studies should fix inputs and acceptance checks,
separating person-hours, blocked waiting, review/rework, and machine cost.
Labor and calendar savings remain to be measured.

\paragraph{Study limitations.}
Imaging studies share one host, domain, and small sets of participating sites.
Our eligibility ledgers and failure taxonomy lack independent adjudication
or inter-rater agreement~\cite{cemri2025mast}. No matched human-effort or
agent-versus-script baseline isolates agency's value. The incomplete
heterogeneous checkpoint needs repeated paired replays and ablations.
Table~\ref{tab:experiment-configurations} records available model and harness
settings and marks historical gaps. Primary-run hyperparameters and compute
accounting remain incomplete. The contribution is federation formation
before conventional FL, rather than a new aggregation or agent-training method.

\paragraph{Clinical validity.}
Tabular patient cohorts were not held out; manual rules are not an external clinical
oracle, and omitted rows prevent full replay. Unresolved OS origins, source
overlap, and censoring mean histogram and curve agreement cannot validate
clinical pooling.

\paragraph{Deployment and privacy.}
Artifact checks do not ensure semantic correctness; hosted-provider access
requires authorization, and workspaces are not sandboxes.
NVIDIA Open Agent Safety Platform offers complementary sandboxing and
independent monitoring~\cite{myers2026agentsafety}. Integration with
\system{} remains future work.
KM histograms are plaintext; deployment requires disclosure-reviewed release
rules, confinement, and operator approval. We claim no formal privacy,
malicious-host resistance, dataset-wide cleaning, or autonomous clinical use.

\section{Conclusion}
\label{sec:conclusion}
FedReady separates agent proposals from runtime-owned artifact acceptance.
Candidate site discovery prioritizes recall, while local verification
establishes task eligibility. Site admission precedes ordinary training or
analysis on fixed verified inputs.
Across imaging and clinical analysis, the resulting workflows connect
task-specific site selection to executable data preparation and retain
evidence for acceptance decisions. Shared guidance extends this design
toward more repeatable preparation by heterogeneous agents.
\label{page:bodyend}

\bibliographystyle{plainnat}
\bibliography{references}

\section*{AI Usage Disclosure}
The AI agents evaluated as part of FedReady are described in the methods and
experimental sections. During preparation of this work, OpenAI Codex assisted
with ideation, literature retrieval and comparison, experiment planning, code
generation and debugging, result summarization, and manuscript drafting,
revision, and LaTeX formatting. Anthropic Claude assisted with manuscript
restructuring and editing, and the overview figure was adapted from a
Claude-assisted design. The authors retain responsibility for the accuracy,
originality, and integrity of the submitted work.

\clearpage
\appendix
\twocolumn[{
\section*{\centering Appendix}
\pdfbookmark[0]{Appendix}{appendix-start}
\vspace{0.5em}
}]
\section{Deployment conditions, safety, and broader impact}
\label{sec:impact}

\paragraph{Scope of the stated properties.}
The main paper's artifact-acceptance property is structural and
conditional on intact acceptance interfaces, not machine-checked. An honest
host is not necessarily bug-free: WF-004 accidentally lost approved contract
fields during synthetic test data construction. This motivates testing the entire
contract-to-test-data-to-executor path; the property alone does not establish that
every host transformation implements its intended interface.
The authority-placement proposition assumes an active site set that only shrinks, no re-entry after exclusion, and the existence of a later stage whose view contains the evidence for the deferred predicate and whose implementation of that predicate is sound and complete.
Dropping the evidence or verifier assumption makes the ``correctable'' half
vacuous. The eligible set is fixed within the modeled episode. Re-admission
would relax the unrecoverability conclusion; versioned re-admission could
remain auditable, but is not evaluated here.

\paragraph{Recovery after changed evidence.}
Exclusion is not permanent across studies. Source correction or newly
available evidence calls for fresh formation checks in a new session, which
may reuse the same contract if its meaning is unchanged. The old outcome is
not rewritten; its source bindings and participating sites describe the earlier episode.
The current bounded retries address implementation failure under an issued
interpretation. They are not an evaluated mechanism for incrementally
reopening screened sites within an existing formation episode.

\paragraph{Security of generated code.}
Generated adapters and trainers run in logical, request-scoped workspaces that constrain paths and imports but are not security sandboxes.
A deployment would additionally require OS- or container-level confinement,
signed and reproducible artifacts, operator approval before generated code is
deployed, and consortium governance of the evidence policy. The evaluated
simulation skips human approval.
The threat model excludes malicious hosts. WF-004 is an accidental host
implementation failure, not evidence of resistance to malicious mutation.

\paragraph{Disclosure.}
Schema-bounded release is a coarse flow policy, not a privacy guarantee.
Aggregate counts, split sizes, reason codes, timing, and the mere fact of a \texttt{screened\_out} response are all side channels, and repeated queries against a stable population compose.
Protecting per-client messages and protecting released aggregates are separate
requirements. Secure aggregation may hide individual contributions under its
assumptions, but does not itself protect a disclosed curve or a single-client
endpoint. Noise would require a specified privacy mechanism and accounting;
neither protection is implemented here, and we make no differential-privacy claim.

\paragraph{Coding-provider boundary.}
Launching a coding worker at a site does not run its language model there.
The Codex backend and the locally hosted VLM are separate channels. The
coding provider must be treated as a recipient of prompts and tool-visible
content, including any local documentation, paths, code, record excerpts,
or diagnostics read into context. The evaluated policy bounds cross-site
FL messages; provider-bound traffic requires its own audit and authorization.
No complete provider-bound payload audit was performed.
All evaluated source data are public. Before private-data deployment, sites
would need to approve provider access or use an on-premises coding model,
enforce tool/network confinement, and audit outgoing payloads. Backend
replacement would need fresh performance evaluation.

\paragraph{Medical use and data.}
The study uses public research datasets, selected following the retinal list of~\cite{saha2026fedagentbench} and downloaded from their original or mirrored public hosts, then registered as simulated sites; all source licenses remain in force and no clinical decision is made anywhere in the pipeline.
Automated site admission carries a specific risk: a site with a bad or ambiguous codebook can be admitted on the strength of a locally plausible interpretation, and a site with an unusual acquisition style can be excluded for reasons that are correct under the contract but undesirable epidemiologically.
The protocol mitigates neither automatically. It instead makes both visible: every exclusion carries a typed reason code and every site admission carries an adapter digest and manifest. It performs no autonomous relabeling, deletion, or repair of site data.
Lower setup friction is the intended benefit and also the main hazard: it becomes cheaper to stand up a federation with the wrong participating sites. Checked artifact acceptance does not prevent that outcome; it makes it auditable after the fact.

\section{Reproducibility and implementation details}
\label{sec:repro}

The prototype is implemented in Python on NVIDIA FLARE.
Live agent requests use a shared typed interface carrying role, action, expected schema, context, site identity, and request identifier.
The primary four-workload snapshot used the hosted Codex configuration for server planning and task-code generation and for site-agent adapter coding; adapter workers were launched at the site boundary with policy-bounded access to the corresponding data directory. Qwen3-VL-8B-Instruct was served through a client-local vLLM-compatible endpoint for bounded visual checks during local adaptation.
The client runtime rejects non-local endpoints for dedicated VLM requests.
Only typed verification summaries cross that FLARE boundary; coding-provider
access is a separate channel described above.
Training packages use PyTorch 2.x and MONAI 1.4+ interfaces where appropriate.

\subsection{Experiment configurations and artifact availability}
\label{sec:experiment-configurations}

Table~\ref{tab:experiment-configurations} separates the reported campaigns.
\textbf{NR} means not recovered from the retained run record or supplied
with that result; a current default is not substituted. Model strings are
the recorded request identifiers, not immutable provider weight snapshots.

\begin{table*}[t]
\caption{Reported experiment configurations. Attempt limits concern local
preparation unless stated otherwise; they are separate from downstream
training rounds and provider/transport retries. NR entries remain historical
reproducibility gaps. Artifact keys are defined below.}
\label{tab:experiment-configurations}
\centering
\small
\setlength{\tabcolsep}{4pt}
\begin{tabular}{@{}>{\raggedright\arraybackslash}p{0.15\textwidth}>{\raggedright\arraybackslash}p{0.29\textwidth}>{\raggedright\arraybackslash}p{0.32\textwidth}>{\raggedright\arraybackslash}p{0.18\textwidth}@{}}
\toprule
Campaign & Model, harness, and reasoning & Budgets and retries & Retained artifacts \\
\midrule
Historical imaging & Codex; model ID, CLI version, and reasoning NR. Local VLM: Qwen3-VL-8B-Instruct; serving version NR. & Four 100-round jobs. Agent deadlines, attempt limits, and request retries NR. & [I]: materialization traces, training plots, failure index; full run environments incomplete. \\
Classification ablations & \nolinkurl{gpt-6-sol}/HIGH through Codex; CLI version NR. & Control/semantic-check pair: one adapter attempt, zero local retries, 1,200-s request deadline, no shared guidance. One run per arm. Discovery-arm local budget and provider retries NR. & Author-supplied matched protocol and token totals; frozen run bundle not supplied. \\
Heterogeneous checkpoint & Sol server; 20 \nolinkurl{gpt-6-sol}/HIGH and 19 \nolinkurl{gpt-6-luna}/LOW site agents; Codex CLI 0.162.0. & One initial adapter request per site plus at most one visual revision; up to five paired guidance waves. Request deadline, transport/provider retries, and repetitions NR. & Author-confirmed settings; [H] design/configuration records. Complete run bundle not supplied. \\
Luna-only & Luna/LOW through Codex; run-linked model ID and CLI version NR. & 39 sites per task, no shared guidance. Deadlines, attempts, retries, and repetitions NR. & Author-supplied task-level outcomes; complete run bundle not supplied. \\
Pi/Nemotron & \nolinkurl{nvidia/nvidia/nemotron-3.5-super-text-preview}/HIGH; Pi 0.85.0. & 39 sites per task. Deadlines, attempts, retries, and repetitions NR. & Author-supplied model pin, harness version, and outcomes; complete four-task bundle not supplied. \\
Clinical survival & \nolinkurl{gpt-5.6-sol}/HIGH; Codex CLI 0.159.2; three isolated workers. & At most five corrective cycles and five analysis attempts; one cycle and two analysis attempts used. Preparation requests: 3,600 s; analysis deadline and provider retries NR. & [S]: fresh-run manifest, selector, prompts, accepted reviews, generated KM code, verification, and reference curves. \\
\bottomrule
\end{tabular}
\end{table*}

\paragraph{Configuration provenance.}
The heterogeneous checkpoint's model IDs, CLI version, initial-request and
visual-revision limits, and five-wave cap are author-confirmed for the
reported mixed study. Visual revision applies to spatial tasks; classification
has no visual-review stage. Initial adapter requests and paired guidance
waves are distinct budget units, not a single total-attempt limit.
The separate Luna-only study's run-linked model ID, CLI version, and budgets
remain NR. Earlier assignments and later prepared protocols in [H] are not
substituted for missing run settings. The classification-ablation model and
budgets were supplied separately with its matched protocol.

\paragraph{Artifact locations.}
[I] is the
\href{https://github.com/ZiyueXu77/AgenticFL/tree/e7c98350300b5fc2a29ceadb6cce7b55794b2f23/docs}{imaging documentation snapshot}
(commit \texttt{e7c9835}), including
\nolinkurl{data-materialization-trace.md},
\nolinkurl{workflow-failure-modes.md}, and training plots under
\texttt{assets/}.
[H] is the
\href{https://github.com/ZiyueXu77/AgenticFL/tree/4d8aa77c77cb59954e3b0649aaaec18dc3257597}{heterogeneous implementation snapshot}
(commit \texttt{4d8aa77}); see
\nolinkurl{configs/corrected_runtime.json},
\nolinkurl{experiments/client_workflowir_transfer_39/}, and
\nolinkurl{experiments/fresh_full39/}.
[S] is the
\href{https://github.com/ZiyueXu77/AgenticFL/tree/7ac0ec21fa610aeb2c840df29c726990b696adfb/experiments/biobank}{survival evidence snapshot}
(commit \texttt{7ac0ec2}); the final manifest is
\nolinkurl{results/fedready_screened_shared_v3_fresh_02/fresh_run_version.json}.
These links identify retained materials, not complete raw-run releases.

The final data snapshot is identified as \texttt{data\_20260731\_1234}; the four 100-round longevity jobs began on August 3, 2026.
Every run is expected to retain: task and profile requests; schema and redaction decisions; agent action records; generated source; source and artifact digests; adapter manifests; sample adapter-check and full-materialization reports; verification dispositions; approved extraction summaries; task-code execution-test logs and repairs; exported job configuration; per-round metrics; and final model state.
The archive-audit script enumerates these artifacts without relying on keyword counts as incident labels.
This is the retention target, not a claim that every artifact was released.
Exact historical replay additionally requires runtime and environment locks,
bridge/prompt digests, generated packages, and source/output bindings.
Table~\ref{tab:experiment-configurations} identifies the available materials;
Section~\ref{sec:tabular-reproduction} details aggregate survival replay.

\subsection{Classification preparation ablations}
The main paper's three-row comparison uses three fresh runs in total: one
matched full-workflow control and one run for each ablation, not three
repetitions per configuration. The reference contains ten eligible sites;
the control and semantic-check ablation share 19 local-preparation candidates,
agents, original inputs and documentation, task instructions, and local
budgets. Each has one adapter attempt and no retries. Agent semantic
reasoning and abstention remain enabled.

The semantic-check ablation removes runtime semantic-evidence presence,
task/meaning consistency, evidence-file provenance, record-to-evidence
binding, codebook/polarity consistency, and class-coverage enforcement.
Schema, execution, full materialization, image readability, uniqueness,
path containment, integer-label checks, and separate original label-source
provenance checks remain enabled. Its one eligible-site exclusion was
provenance-related; the available result does not explain its causal
relationship to the removed gate.

Standalone end-to-end token totals are 10,467,482 for the matched control,
9,359,084 without semantic-evidence and class-coverage checks, and 11,500,395
without candidate site discovery. Relative to control, these are 1.000,
0.894, and 1.099, respectively. The semantic-check ablation uses 10.6\% fewer
tokens but excludes one eligible site, so it does not provide equivalent
formation outcomes. The discovery ablation retains the same ten sites with
9.9\% more tokens. These are single-run observations, not repeated-run cost
estimates. Per-arm implementation-failure and elapsed-time accounting were
not reported; missing values are not zeros.

\subsection{Standalone agent preparation studies}
\label{sec:pi-nemotron}
The Luna-only study assigns Luna-low site agents through the Codex harness
to the same 39 sites, without shared guidance. It misses five eligible
site--task pairs and reports zero incorrect admissions: one miss each in
disc segmentation, cup segmentation, and disc detection, and two in
glaucoma classification.
These are final false negatives against the task reference; their
classification as evidence-related exclusions or implementation failures
was not supplied. They must not be conflated with justified screening of
ineligible sites or with the separate heterogeneous-guidance checkpoint.

The Nemotron study uses Pi with Nemotron 3.5 Super/HIGH
(Table~\ref{tab:experiment-configurations}). Its preparation outcomes and the
Luna-only outcomes are separate from the heterogeneous-agent checkpoint and
the four downstream-training runs.

\begin{table*}[t]
\caption{Pi with Nemotron 3.5 Super/HIGH: preparation outcomes against the
task-eligibility reference. Implementation failures concern eligible sites,
not evidence that those sites lack supported data.}
\label{tab:pi-nemotron}
\centering
\small
\begin{tabular}{@{}lrrrr@{}}
\toprule
Task & Eligible & Ready & Implementation failures & Incorrect admissions \\
\midrule
Disc segmentation & 8 & 8 & 0 & 0 \\
Cup segmentation & 6 & 4 & 2 & 0 \\
Glaucoma classification & 10 & 9 & 1 & 0 \\
Disc detection & 8 & 6 & 2 & 0 \\
\midrule
Total & 32 & 27 & 5 & 0 \\
\bottomrule
\end{tabular}
\end{table*}

The Luna and Nemotron studies share the eligibility reference. Their missed
eligible sites are disjoint within each task: no site missed by Luna is also
missed by Nemotron for that task. Whether their complementary misses support
an effective ensemble or fallback remains a separate experiment.

\section{Site eligibility evidence policy}
\label{sec:evidence-policy}

Candidate site discovery finds potentially suitable sites from approved
profiles. Local eligibility verification checks whether each candidate
supports the task using local evidence and prepared artifacts. Site admission
accepts verified sites into the federation. The requirements below govern
local eligibility verification, not discovery alone.

Optic-disc and optic-cup segmentation require locally provable spatial supervision for the requested structure and a readable compatible input image.
Contour coordinates, binary masks, soft maps, and documented multi-value masks are eligible if a local adapter can convert them without inventing semantics; the main paper's local-translation figure contrasts these encodings with the required binary disc mask.
Glaucoma classification requires explicit local evidence mapping records to glaucoma or referable-glaucoma status; structural proxies, undocumented numeric codes, abbreviations without local definition, or dataset-name memory are insufficient.
Detection requires the disc-eligible source plus a generated contract that defines normalized or pixel-coordinate boxes consistently with the generated trainer.

The reference ledgers are evaluator artifacts, not live allow-lists.
The primary run admits 8 disc sites, 6 cup sites, 10 classification sites, and 8 detection sites.
Because one physical package can be eligible for multiple tasks, the 32 final site-task memberships should not be interpreted as 32 distinct organizations.

\section{Failure analysis and complete stable-workflow index}
\label{sec:failures}

\subsection{Archive scope and evidence criteria}

We run a read-only audit over all retained development artifacts through August 11, 2026; coverage is exhaustive within that archive, not over all attempted executions.
The audit inventories 254,854 files in 87 candidate workflow, test, or pilot roots; analyzes all 89,113 text/structured artifacts (626.7 MB); parses 153,672 JSON documents or JSONL rows; and reconciles incidents with code history, retry artifacts, or contemporaneous diagnosis.
An incident enters the taxonomy only when a real run shows a causal artifact chain.
Evidence grade E1 additionally requires a validating retry or regression test; E2 requires a timestamped run plus code or documented diagnosis.
Constructed red-team cases and unobserved invariants are excluded.

The full archive contains 23 evidence-graded incidents (10 E1, 13 E2).
Four are specific to the experimental sample-quality screening that this paper places out of scope: WF-002, WF-011, WF-012, and WF-013, each a visual-QC verdict on image quality rather than task compatibility. We exclude them and analyze the remaining 19 stable-workflow incidents: 8 E1 cases with a validating retry or regression and 11 E2 cases with timestamped evidence and diagnosis.

\paragraph{Partition by what was violated.}
Classifying each incident by what it violated rather than by the component that surfaced it gives four disjoint classes.

\emph{Authority}, I2 (5): a role decided outside its ownership, or a self-reported success was consumed as a verifier result: WF-003 and WF-008 (generated materializers inventing undeclared invariants), WF-009 (an adapter inventing label semantics from pixel values), WF-010 and WF-023 (backend outcomes consumed as data outcomes).

\emph{Scope}, I1 and I3 (5): a fact changed state at a granularity, or through a channel, that its evidence does not license: WF-001 (a path token deciding record semantics) and WF-007 (redaction removing evidence needed for candidate site discovery) violate I1; WF-006 (a log line overriding completion evidence), WF-017 (one client rejection aborting a run), and WF-018 (a reporting call overriding runtime facts) violate I3.

\emph{State loss} (5): a negotiated or accepted artifact was mutated, re-encoded, or resolved against the wrong base between acceptance and use: WF-004, WF-005, WF-014, WF-015, WF-021.

\emph{Environment} (4): a defect that surfaced only when the exact deployed code and runtime executed: WF-016, WF-019, WF-020, WF-022.

We assign 15 of 19 to the first three classes. Better generation might reduce
incidents in any class; this partition cannot isolate model quality from
interface design. Several incidents span classes, and we assign the earliest
identified violation. This is author coding without a blinded second
annotator, not an independently validated taxonomy or an inter-rater study
such as MAST~\cite{cemri2025mast}.

The 87 archive roots, artifact counts, retries, and repeated logs are not
independent exposure units. No reliable denominator of all workflow runs,
site-task attempts, or boundary transitions is reconstructed, so 15/19 is a
share of retained incident types, not a failure probability or an estimated
repair benefit. The primary imaging snapshot uses one coding-agent family;
unmatched OpenHands stress traces do not establish backend generality.
Validation would require a frozen codebook, independent annotation with
agreement and adjudication, deduplicated run/attempt exposure accounting,
and prospective held-out traces across backend families.

\subsection{Selected causal failures}

Table~\ref{tab:failures} shows the E1 cases with the clearest cross-boundary consequence.
Silent failures are especially important because every component may report local success while collective state is wrong.

\begin{table*}[t]
\caption{Selected E1 coordination failures. Each occurred in an executable workflow and has a validating retry or regression test.}
\label{tab:failures}
\centering
\small
\begin{tabular}{@{}>{\raggedright\arraybackslash}p{0.075\textwidth}>{\raggedright\arraybackslash}p{0.25\textwidth}>{\raggedright\arraybackslash}p{0.27\textwidth}>{\raggedright\arraybackslash}p{0.31\textwidth}@{}}
\toprule
ID & Boundary failure & Consequence at occurrence & Protocol-level repair \\
\midrule
WF-001 & Weak path token treated as record-level evidence, then used in candidate site discovery & \textbf{Silent}: one eligible site and 4,217 records omitted from a completed run & Narrow observation ownership; typed replay recovered and verified all 4,217 records \\
WF-003/008 & Generated materializer invented invariants not declared by the contract & 1,949 valid records across two sites dropped after successful adapter work & Contract-declared invariants became the complete authority; focused retries recovered both sites \\
WF-004 & Host synthetic test data constructor erased fields from an approved generated contract & Seven coding-agent revisions blamed the trainer; the one-client simulation was never reached & Preserve the negotiated contract end to end and validate a complete synthetic test record \\
WF-005 & Shell-valid JSON fragmented by the deployed FL runner's tokenizer & Valid manifests appeared to lack required fields despite static validation & Validate compact arguments with the actual deployed tokenizer \\
WF-016 & Static checks missed CLI, device, update-delta, and installed-API defects & Four runtime incompatibilities; generated package unusable & Execute the exact exported interface in the one-client simulation; the third runtime revision succeeded \\
WF-006 & Any \texttt{ERROR} line overrode stronger completion evidence & A valid 100-round run was labeled failed and an unnecessary rerun started & Model, metric, and controller terminal state now govern; recoverable events remain warnings \\
\bottomrule
\end{tabular}
\end{table*}

\paragraph{WF-001: lossy candidate site discovery silently changed the population.}
A directory \texttt{seg\_dataset/\{glaucoma,normal\}} contained 4,217 classification images.
A structural parser interpreted the weak token ``seg'' as mask evidence and omitted the class-folder evidence from the shareable profile.
The round-1 agent reasonably returned ``not applicable'' (a site-screening outcome), the client never reached its local adapter, and the workflow reported success.
After narrowing the parser's ownership rule, a one-client replay produced a verified 4,217-record manifest.
The agent could not recover evidence it was never shown; the fault belonged to the candidate site discovery interface.

\paragraph{WF-004: host state misdirected the repair loop.}
The agent-generated detection trainer included the required contract fields, but host-side synthetic test data construction discarded them.
Seven revisions could not repair the host's corrupted view.
Preserving the full approved contract enabled task-code execution tests and later training.
Together these cases support a governance rule: feedback must preserve causal ownership, or the workflow spends its repair budget on the wrong component.

\paragraph{Backend stress behavior.}
Unmatched OpenHands/Qwen stress runs (SDK 1.27.0) either produced no validated
datalist or failed before the required training-code specification; the
runtime withheld job export. These cases illustrate the intended fail-closed
behavior: backend failure reduces availability rather than replacing
executable evidence with self-report. They are not a matched model comparison.

\subsection{Complete retained index}

Table~\ref{tab:allfailures} gives a compact summary of the 19 retained incidents.
E1 means a timestamped trace plus a validating retry or regression test; E2 means a timestamped trace and diagnosis without a clean controlled validation.

\begin{table*}[t]
\caption{Descriptive index of retained stable-workflow incidents, derived from the versioned engineering archive.}
\label{tab:allfailures}
\centering
\small
\begin{tabular}{@{}llp{0.71\textwidth}@{}}
\toprule
ID & Grade & Observed failure \\
\midrule
WF-001 & E1 & Weak path token erased classification evidence and silently excluded 4,217 valid records. \\
WF-003 & E1 & A generated materializer invented an undeclared pixel-uniqueness invariant and dropped a valid site. \\
WF-004 & E1 & Host synthetic test data construction erased approved contract fields, causing seven misdirected coding-agent revisions. \\
WF-005 & E1 & Shell-valid JSON was fragmented by the deployed runner tokenizer. \\
WF-006 & E1 & A recoverable communication event overrode stronger evidence of completed 100-round training. \\
WF-007 & E2 & Privacy hardening redacted target vocabulary required by candidate site discovery, reducing participating sites for all tasks. \\
WF-008 & E1 & Sample adapter checks passed, but a generated full-data nesting invariant later removed 1,200 records. \\
WF-009 & E2 & An adapter inferred confidence semantics from grayscale values without local metadata. \\
WF-010 & E2 & Backend request completion masked an unimplemented adapter skeleton and resembled data infeasibility. \\
WF-014 & E1 & Adapter-input encoding and materialized-output encoding were conflated. \\
WF-015 & E2 & A generated package path alias was resolved against the wrong workspace base. \\
WF-016 & E1 & Static checks missed four exact CLI/device/update/API runtime incompatibilities. \\
WF-017 & E2 & Strict quorum amplified one client readiness rejection into a whole-run abort. \\
WF-018 & E2 & A free-form reporting call obscured completed deterministic extraction facts. \\
WF-019 & E2 & A syntax error in staged runtime code prevented client startup. \\
WF-020 & E2 & Multiple guardrail-worker failures escalated into a fatal system error. \\
WF-021 & E2 & Resume schemas and APIs drifted, producing three control-plane aborts. \\
WF-022 & E2 & A live run launched without the required agent executable in the runtime environment. \\
WF-023 & E2 & Backend iteration and provider failures could masquerade as client/data outcomes. \\
\bottomrule
\end{tabular}
\end{table*}

\section{Message and artifact ledger}
\label{sec:ledger}

Table~\ref{tab:ledger} expands the main-text information-flow description into concrete producer-consumer interfaces.
The decisive distinction is between a message that may cross the runtime boundary and a private artifact that is consumed by a local verifier.

\begin{table*}[t]
\caption{Key interfaces in the implemented workflow. Server-to-client and client-to-server denote FL-runtime transport; ``local'' artifacts never cross that boundary.}
\label{tab:ledger}
\centering
\small
\setlength{\tabcolsep}{3pt}
\begin{tabular}{@{}>{\raggedright\arraybackslash}p{0.19\textwidth}>{\raggedright\arraybackslash}p{0.19\textwidth}>{\raggedright\arraybackslash}p{0.25\textwidth}>{\raggedright\arraybackslash}p{0.31\textwidth}@{}}
\toprule
Artifact or message & Producer and consumer & Information carried & Check before state effect \\
\midrule
Client inquiry (v1) & Controller to client executor (server-to-client) & Task intent and bounded candidate site discovery instructions & Header binding, allow-listed action/phase, client policy authorization \\
Client response (v1) & Client executor to controller (client-to-server) & Redacted aggregate profile, applicability, counts, privacy metadata & Schema and redaction checks; exact session, correlation, task, site, and payload-identity binding \\
Site extraction policy (v1) & Server decision flow to selected client & Locked contract, site policy, adapter rule, strategy digest, optional materializer source and digest & Server disclosure filter; client action/phase/schema authorization; contract and materializer integrity \\
Local adapter manifest (v1) & Site agent to adapter runtime (local) & Candidate adapter entry point and private record bindings & Pre-agent source snapshot, compilation, sample adapter checks, provenance, schema, and source-digest checks \\
Local visual verdict (v1) & Local visual witness to client runtime (local) & Reference-conditioned finding for one raw input and optional contract-owned label rendering & Localhost-only endpoint, reference digest, bounded issue schema, concordance/fallback policy; never sufficient alone for site admission \\
Runtime verification record (v1) & Adapter runtime to client executor (local) & Harness-owned executable verdict bound to exact adapter source & Required local checks complete; agent-reported status is overwritten by runtime facts \\
Client extraction response (v1) & Client executor to controller (client-to-server) & Status/reason, aggregate split counts, verifier outcomes, contract/policy digests, redaction declaration & Outgoing disclosure-filter/schema checks plus reply binding; coherent \emph{extracted} results are required before site admission \\
Training plan / code spec (v1) & Server planner/coder to exporter (server local) & Approved data contract, synthetic test data interface, contained trainer package & Schema, path, containment, and compilation checks plus task-code execution tests in a runtime-owned one-client simulation \\
Client training config (v1) & Exporter to admitted client launcher (job deployment) & Fixed contract, participating-site list, artifact digests, and expected local manifest & Launcher resolves local prepared data and validates schema, splits, verification, and digest bindings before task code \\
FL model update & FedAvg server and client trainer & Global parameters outward; parameter differences, sample count, and aggregate metrics inward & No agent call; update type and weight checks, controller completion, model persistence, and metric artifact \\
Audit event & Runtime component to audit graph (boundary-local or server) & Phase, role/channel/action, decision, correlation metadata, reason, payload/artifact digests & Append-only event identity and digest linkage; private artifact bodies remain at the owning site \\
\bottomrule
\end{tabular}
\end{table*}

\section{Tabular survival preparation: evidence, verification, and reproducibility}
\label{sec:tabular-details}

\subsection{Sources, execution, and scope}

The experiment is preserved in the
\href{https://github.com/ZiyueXu77/AgenticFL/tree/7ac0ec21fa610aeb2c840df29c726990b696adfb/experiments/biobank}{versioned biobank repository snapshot}
(commit \texttt{7ac0ec2}). We compare the final screened fresh run,
\nolinkurl{fedready_screened_shared_v3_fresh_02}, with the separately prepared
manual V5 reference, \nolinkurl{km_manual_v5_20261008_v1}.
Their retained curves and reports are under \texttt{results/}; the
\href{https://github.com/ZiyueXu77/AgenticFL/blob/7ac0ec21fa610aeb2c840df29c726990b696adfb/experiments/biobank/results/compare_curves.py}{offline comparison script}
reproduces aggregate agreement without patient inputs or agent calls.
The source counts are 290, 177, 55, 96,
62, 130, 585, 161, 461, and 172 for GSE14333, GSE17536, GSE17537, GSE33113,
GSE35896, GSE37892, GSE39582, Ethiopia, TCGA\_COAD, and TCGA\_READ,
respectively. Public source patient cohorts simulate private clients on one host.

The run uses the Codex backend with three workers. Disposable site-only
workspaces and
checksum-bound input staging separate practical access scopes; they do not
provide OS-level isolation. Adapters, patient rows, and record crosswalks are
excluded from cross-site FL messages, not guaranteed hidden from the hosted
coding provider. The shared-host numerical
checker reads accepted local outputs for offline verification. Plaintext
histograms, summaries, and counts provide no differential-privacy or
cryptographic guarantee.

The 98-file selector binds clinical metadata, original publications,
dictionaries, and mechanically extracted reading copies. Gene-expression
inputs, manual mappings and policy files, prepared tables, and previous agent
artifacts are excluded. Fresh execution regenerates summaries, contracts,
adapters, and decisions, but these patient cohorts were used in earlier prompt and
reference development. This is a fresh-execution test, not an untouched
held-out benchmark. The manual implementation was separately written and
documented, but provides neither independent clinical adjudication nor an
unquestionable oracle.

\subsection{Readiness decisions beyond a shared schema}

\paragraph{Relational identity.}
GDC files contain cases with nested diagnoses and follow-ups. The adapter must
emit the agreed case-level unit rather than treating every nested object as an
independent participant. Source IDs, join cardinality, and retained output
identity require local checks and crosswalks. For GEO, sample accessions can
support mechanical linkage without independently verifying unique patients.
In GSE35896, all 62 secondary clinical IDs match GEO samples, but this join
neither authenticates the clinical mirror nor resolves conflicting attributes.
Cross-client patient overlap is not established by within-client uniqueness.

\paragraph{Temporal and endpoint semantics.}
Overall survival, disease-free survival, and colorectal-cancer-specific
survival remain separate endpoint families. TCGA OS combines vital status with
death or follow-up durations in the selected case/diagnosis context. Native
months are converted to years only where documentation supports the unit.
Numerical ranges, coherent date ordering, and plausible distributions do not
independently define units, event polarity, censoring, death handling, or time
origin. Ethiopia's event definition counts CRC death, while censoring does not
imply that a patient is known to be alive; a chronology contradiction remains held. The
workflow does not replace documented duration fields with a guessed
date-derived target or clip contradictory durations into validity.

\paragraph{Population and selection.}
Eligibility and endpoint applicability precede missing-target handling.
Normal tissue, ambiguous primary origin, prior treatment, and subgroup
restrictions are accounted for explicitly. In GSE14333, primary DFS uses
185 Dukes B/C records with 48 recurrences. The manual reference preserves
226 A/B/C records with 50 events as a separate sensitivity population;
41 complete stage-A pairs are not silently included in primary analysis.
For GSE37892, including only records with recorded metastasis would select
on the outcome, so complete arithmetic for that subset is insufficient to
justify a survival estimate for the full patient cohort.

\paragraph{Field-level uncertainty.}
A supported survival target can coexist with an unsupported covariate.
Disputed sex codes, unknown staging systems, and ambiguous age reference
points retain native values and uncertainty rather than forcing a crosswalk.
Dukes stages are not silently converted to AJCC stages; AJCC/TNM edition and source context
are retained where available. GDC age-at-index is not automatically
age-at-diagnosis, and unknown obfuscation flags remain unknown. A disagreement
in one covariate need not exclude otherwise supported endpoints, whereas
complete covariates cannot rescue an undefined endpoint.

\subsection{Native columns and evidence used to reconcile them}
\label{sec:tabular-columns}

Table~\ref{tab:tabular-native-columns} preserves source field spellings from
the branch's native-column inventory and manual builders. GEO names denote
keys inside \nolinkurl{!Sample_characteristics_ch1}, not necessarily physical
CSV headers; \nolinkurl{^SAMPLE} supplies the sample accession. GSE14333 packs
several keys into one semicolon-delimited characteristic. Parsers can
case-normalize keys internally (for example, \texttt{DFS\_Time} becomes
\texttt{dfs\_time}) without changing the deposited field's identity. GDC names
denote nested JSON paths. The early inventory supplies spelling only;
final decisions follow the screened contract and manual V5 policy.

\begin{table*}[t]
\caption{Native target fields and decision context. Field presence alone does not establish readiness; the following paragraphs identify supporting evidence and unresolved assumptions.}
\label{tab:tabular-native-columns}
\centering
\small
\setlength{\tabcolsep}{4pt}
\renewcommand{\arraystretch}{1.15}
\begin{tabular}{@{}>{\raggedright\arraybackslash}p{0.15\textwidth}>{\raggedright\arraybackslash}p{0.39\textwidth}>{\raggedright\arraybackslash}p{0.40\textwidth}@{}}
\toprule
Source patient cohort & Target or candidate-target fields & Identity, population, and covariate fields \\
\midrule
GSE14333 &
\nolinkurl{DFS_Time}, \nolinkurl{DFS_Cens} &
\nolinkurl{^SAMPLE}, \nolinkurl{DukesStage}, \nolinkurl{Age_Diag}, \nolinkurl{Gender}, \nolinkurl{Location}. Packed characteristic parsing. \\
GSE17536 and GSE17537 &
\texttt{overall survival follow-up time};
\texttt{overall\_event (death from any cause)} &
\nolinkurl{^SAMPLE}, \texttt{!Sample\_title};
\nolinkurl{age}, \nolinkurl{gender}, \nolinkurl{ajcc_stage}, \nolinkurl{grade}.
Other DFS/DSS fields are not OS aliases. \\
GSE33113 &
\texttt{meta or recurrence within 3 years};
\texttt{time to meta or recurrence} &
\nolinkurl{^SAMPLE}, \nolinkurl{tissue};
\texttt{age at diagnosis}, \nolinkurl{Sex}, \texttt{disease status}.
Tumors and normal mucosa require different dispositions. \\
GSE35896 &
\texttt{DFS (in months)}, \texttt{DFS event} in a secondary clinical CSV &
CSV \nolinkurl{ID_REF} joined to GEO \nolinkurl{^SAMPLE};
\texttt{Age (in years)}, \nolinkurl{Gender}, \texttt{Dukes Stage}, \nolinkurl{Location};
direct GEO \nolinkurl{microsatellite.status} and sex metadata. \\
GSE37892 &
\texttt{date at surgery}, \texttt{date at distant metastasis},
\texttt{date at last contact} &
\nolinkurl{^SAMPLE}; \texttt{age at diagnosis}, \nolinkurl{gender},
\nolinkurl{Stage}, \nolinkurl{localisation}. Date-derived intervals are candidates only. \\
GSE39582 &
\texttt{os.delay (months)}, \nolinkurl{os.event} &
\nolinkurl{dataset}, \nolinkurl{tnm.stage}, \nolinkurl{tnm.m};
\texttt{age.at.diagnosis (year)}, \nolinkurl{Sex},
\nolinkurl{tumor.location}, \nolinkurl{mmr.status}.
RFS remains a different endpoint family. \\
Ethiopia &
\nolinkurl{TIME}, \nolinkurl{VSTATUS}, \nolinkurl{statusfn};
\nolinkurl{Date_Dx}, \nolinkurl{Date_death} for chronology &
Immutable CSV row position; \nolinkurl{Age}, \nolinkurl{Sex},
\nolinkurl{TNM_stage}, \nolinkurl{Location}, \nolinkurl{Hist_grade}.
\nolinkurl{FU_last} and \nolinkurl{Status_patient} do not replace the issued target. \\
TCGA\_COAD and TCGA\_READ &
\nolinkurl{demographic.vital_status},
\nolinkurl{demographic.days_to_death},
\nolinkurl{diagnoses.days_to_last_follow_up},
\nolinkurl{follow_ups.days_to_follow_up} &
\nolinkurl{submitter_id}, \nolinkurl{primary_site},
\nolinkurl{diagnoses.diagnosis_is_primary_disease},
\nolinkurl{diagnoses.prior_treatment},
\nolinkurl{diagnoses.tissue_or_organ_of_origin};
\nolinkurl{demographic.age_at_index}, \nolinkurl{demographic.age_is_obfuscated},
\nolinkurl{demographic.sex_at_birth}, \nolinkurl{diagnoses.ajcc_pathologic_stage}. \\
\bottomrule
\end{tabular}
\end{table*}

\paragraph{GSE14333: reverse coding and population scope.}
Native \texttt{DFS\_Time}, \texttt{DFS\_Cens}, and \texttt{DukesStage}
are reconciled with the publication's patient/statistical
methods~\cite{jorissen2009metastasis}, recurrence counts, and duration
distributions. For the documented B/C population, the issued mapping is
\texttt{duration\_years=DFS\_Time/12} and \texttt{event=1-DFS\_Cens}.
This is documentary-plus-aggregate corroboration, not a recovered literal
deposited codebook. The contract reconciles 94 stage-B and 91 stage-C records
to 185 ready targets, while 44 stage-A and 61 stage-D records are held.
Its local \texttt{dfs\_transform\_violations} measure must be zero and its
unique-output count must be 290. The manual audit identifies 226 complete
A/B/C pairs, including 41 complete A pairs, and 64 missing native pairs
overall: completeness cannot determine primary eligibility. The publication
and deposit differ by three B/C records; current deposit counts are not
forced to match historical publication counts. Documentary age units and
matching aggregate range/median support \texttt{Age\_Diag} in years.
\texttt{DukesStage} remains Dukes without an assumed AJCC crosswalk.

\paragraph{GSE17536/GSE17537: identity and explicit event wording.}
The event field specifies death from any cause and uses
\texttt{death}/\texttt{no death}. GEO series descriptions, the publication for
these patient cohorts \href{https://pmc.ncbi.nlm.nih.gov/articles/PMC3388775/}{PMC3388775},
and distinct patient-numbered sample titles jointly support 177 Moffitt
and 55 Vanderbilt patients. Native pairs corroborate 177 OS targets with
73 deaths and 55 targets with 20 deaths; documentary follow-up units
support division by 12. Count agreement alone is insufficient, and no
independent patient-key column is available. Their parent series GSE17538
must not be counted as another independent patient cohort. Exact interval origin
and staging edition remain unresolved. Seven compound \texttt{grade}
values in GSE17537 remain raw rather than being silently decoded or used
to discard supported OS targets.

\paragraph{Ethiopia: triangulation without silent repair.}
The CRC-specific publication~\cite{teka2021colorectal} defines the endpoint
and diagnosis origin; the broader data publication
\href{https://pmc.ncbi.nlm.nih.gov/articles/PMC8367784/}{PMC8367784}
supplies a colorectal life table. Native \texttt{VSTATUS}/\texttt{statusfn}
agree across all 161 released records. Their 75 events and the
2,937-person-month total corroborate \texttt{TIME}/\texttt{VSTATUS};
the reference reports reproduction of all 36 published risk/death/censor
cells before exclusion. The final mapping uses \texttt{TIME/12} and
\texttt{VSTATUS=1} for CRC death; \texttt{0} denotes censoring, not
necessarily known-alive status. \texttt{Date\_Dx}/\texttt{Date\_death} reveal
one endpoint date before diagnosis while \texttt{TIME} is zero. That
record remains held, leaving 160 ready records and 74 events; dates do not
overwrite \texttt{TIME}. Documentary disagreement in numeric \texttt{Sex}
coding remains unresolved and raw. Published/native follow-up discrepancies
also remain limitations rather than being erased by matching totals.

\paragraph{TCGA: release-aware paths and overlap bounds.}
Staged GDC dictionary/term definitions and outcome methods~\cite{liu2018clinical}
inform entity roles, day offsets, and OS construction. GDC API release
guidance~\cite{gdcClinicalDocumentation} includes
\nolinkurl{follow_ups.days_to_follow_up}; using only an older follow-up
path can lose later observations. Dead cases use death days; alive cases
use the largest applicable nonnegative contact/follow-up interval.
Days are divided by 365.25; a missing death time is not replaced by
living follow-up. \nolinkurl{diagnosis_is_primary_disease},
\nolinkurl{primary_site}, \nolinkurl{tissue_or_organ_of_origin}, and
\nolinkurl{prior_treatment} guide population screening, while
\texttt{submitter\_id} binds case identity. A rectal biopsy site does not
establish rectal primary origin. Current COAD aggregates contain 461 cases,
458 mechanically complete OS pairs, and three prior-treatment holds.
Because their overlap is initially unavailable, the generated contract
uses a 455--458 ready bound. Local measured reconciliation establishes
that all three holds are target-complete, leaving 455 ready and three
target-missing cases. READ likewise resolves anatomy/treatment/missingness
overlap rather than summing counts blindly. These are issued workflow
rules, not proof that every eligibility boundary matches the reference.
Historical patient-cohort sizes are not current-release quotas. Age-at-index,
obfuscation, AJCC edition, and diagnosis context remain explicit.

\paragraph{GSE39582: endpoint corroboration and localized conflicts.}
Native \texttt{os.delay (months)}/\texttt{os.event} supply explicit units
and candidate OS coding. \texttt{dataset=Non Tumoral} identifies 19
controls among 585 samples. The patient-cohort paper
\href{https://pmc.ncbi.nlm.nih.gov/articles/PMC3660251/}{PMC3660251},
CMS paper \href{https://pmc.ncbi.nlm.nih.gov/articles/PMC4636487/}{PMC4636487},
and outcome paper
\href{https://pmc.ncbi.nlm.nih.gov/articles/PMC8657766/}{PMC8657766}
provide same-source context. The outcome paper's OS table reports
191 events, 371 non-events, and four missing, corroborating 562 complete
tumor pairs before the agent's stage-0 exclusions. The agent contract
holds four \texttt{tnm.stage=0} records as well as controls, leaving
558 ready and four target-missing records. Manual V5 instead retains
stage-0 records for supported OS without normalizing stage;
this policy difference contributes to the OS discrepancy.
The reference also flags \texttt{tnm.stage=2} with \texttt{tnm.m=M1}
as a covariate contradiction. It does not invalidate every supported
OS target or justify replacing OS with RFS. Exact time origin remains
unresolved.

\paragraph{GSE35896: join success without authenticated provenance.}
Mirror \texttt{ID\_REF} matches all 62 GEO accessions and offers
\texttt{DFS (in months)}, \texttt{DFS event}, \texttt{Age (in years)},
and \texttt{Gender}. Primary GEO metadata/publication evidence is compared
with the mirror; header wording alone is not trusted. The archived contract
reports 32 joined-record sex disagreements, incompatible age evidence,
unauthenticated clinical provenance, and one villous-adenoma-only pathology
conflict. It requires an actual post-execution 62-record join count,
exclusion of an index-only export artifact, and null targets on held rows.
Mechanical linkage succeeds but DFS remains unissued. Direct-source
\texttt{microsatellite.status} can remain raw without legitimizing
secondary survival fields.

\paragraph{GSE33113/GSE37892: negative results test abstention.}
GSE33113 \texttt{tissue} separates 90 tumors from six normal specimens.
The initial audit finds 89 complete recurrence pairs and 18
\texttt{yes} records, including time 1278 under a flag labeled
\texttt{within 3 years}. This warns against inferring units from range or
assuming \texttt{no} means event-free censoring. GEO metadata and the
available publication abstract do not resolve all required meanings;
targets stay null. GSE37892's three dates yield 130 mechanically
calculable intervals, including 37 recorded metastases, and chronology
checks succeed. Series/sample metadata and the primary abstract support
patient-cohort scope but do not define absent-date censoring or death handling.
All targets remain held; selecting only the 37 metastases would bias the
population. Aggregate review also challenges normalized \texttt{Stage}
when its staging system is unknown. Unsupported stage and age units remain raw.

\paragraph{Limits of corroboration.}
Documentation justifies candidate meanings; duplicate fields, chronology,
native distributions, joins, and published aggregates corroborate or
contradict them; measured assertions test implementation against issued
rules. None alone is a semantic oracle. The compact archive omits native
rows, full candidate site discovery summaries, and local verifiers, so these examples
are traced to archived policies and reported audits, not newly recomputed
patient-level cross-checks.

\subsection{Contract and verification responsibilities}

The configured task brief identifies output fields and requests named
screening sections: analysis unit, eligibility, evidence precedence,
subgroup applicability, accounting, and covariate handling. These section
names and the clinical schema are task inputs. The server generates their
substance and site rules from local evidence; task-neutral runtime code
checks representation and contract conformance, rather than deciding
clinical meanings from hard-coded patient-cohort names.

\begin{enumerate}[leftmargin=*,itemsep=3pt,topsep=3pt]
\item \textbf{Local evidence and policy proposal.} A site agent inspects
only staged site files and returns source-field descriptions, aggregate
structure, citations, and unresolved questions. Publications are evidence,
not assumed complete column-level codebooks.
\item \textbf{Shared contract.} The server drafts endpoint identities,
duration/event rules, eligibility partitions, field handling, site guidance,
and measurable assertions. A global-contract defect requires a new version
before affected clients rerun.
\item \textbf{Executable local adaptation.} Site agents generate an adapter
and, when contracted, a measurement program. Local scripts process all
records; no per-record LLM inspection is performed. The runner checks
input hashes, ordered output schema, site/endpoint binding, unique nonempty
IDs, finite nonnegative durations, binary events, and recognized
\texttt{ready}, \texttt{held}, or \texttt{missing\_target} statuses.
Nonready rows must have blank targets.
\item \textbf{Execution-derived checks.} Runner-owned counts and
agent-generated source/output measurements are evaluated by fixed
\texttt{eq}, \texttt{ge}, \texttt{le}, or \texttt{between} predicates.
Assertions carry evidence, scope, and a failure disposition
(\texttt{revise\_local} or \texttt{review}). Client-supplied pass flags are
insufficient. Generated measurement programs can still repeat adapter
assumptions, so their execution is not independent semantic verification.
\item \textbf{Aggregate review.} A reviewer sees the contract, evidence
summaries, and measured aggregates, not patient rows or adapter code.
It reconciles population counts, parsing/linkage evidence, and unsupported
normalization, and requests bounded correction. Pre-execution checklist
narratives remain distinct from actual output measurements.
\item \textbf{Analysis on fixed inputs and offline comparison.} A generated
four-file NVFlare application consumes only accepted ready records for the
exact endpoint, binds input and contract digests, and undergoes review,
simulation, and independent numerical checking. The manual-reference
comparison occurs separately after agent execution.
\end{enumerate}

The final run permits at most five corrective cycles and five analysis
attempts. It completed one aggregate corrective cycle: GSE37892 required
resolution of unsupported stage normalization, and TCGA\_COAD required
reconciliation of source/output stage coverage. These were covariate/measurement
issues, not requests to include held endpoints. No global-contract revision
was needed. The analysis completed on attempt two. The summary reports
recovery from a local TCGA\_READ adapter failure; the compact archive omits
the complete local attempt trace.

\subsection{Held endpoints and record accounting}

The endpoint at GSE33113 remains held because interval unit/origin and event/censor semantics
are not sufficiently supported. The endpoint at GSE35896 remains held because a mechanically
complete secondary join does not authenticate provenance or DFS semantics.
The endpoint at GSE37892 remains held because dates alone do not resolve metastasis event
definition, censoring, death handling, origin, and identity. These are
documented dispositions under the experiment's evidence policy, rather than
runtime failures or proof that the patient cohorts can never support analysis.

The final agent tables retain 25 normal controls as held records; the manual
tables omit them while keeping audit decisions. Different accounting
conventions should not be mistaken for different analyzed populations.
Both retain distinct target-missing and held dispositions. The desired
outcome is correct record inclusion and justified holds, not maximum
record count.

\subsection{Numerical checks versus reference agreement}

Accepted durations in years are mapped to month bins
$b=\max(0,\lceil 12t-10^{-8}\rceil)$. Clients produce endpoint-specific event
counts $d_{i,b}$ and censor counts $c_{i,b}$. The server sums these, computes
the initial risk set from their totals, and updates
\[
 \begin{aligned}
 S_b &= S_{b-1}\left(1-\frac{\sum_i d_{i,b}}{r_b}\right),\\
 r_{b+1}&=r_b-\sum_i(d_{i,b}+c_{i,b}).
 \end{aligned}
\]
Events use the risk set before either events or censorings in that bin are
removed. This is a product-limit calculation on the configured monthly
discretization; exact agreement does not refer to unbinned event times.
The reported slot width is one month in both arms. Other widths should be
specified in the analysis contract and compared using the same binning;
only the monthly setting is evaluated here.

The separately implemented checker reconstructs monthly risk sets from the actual
accepted local outputs, validates endpoint and digest binding, and checks
counts and survival values. All three agent analyses report zero maximum
arithmetic error. This tests the generated application on its accepted
population; it does not validate the original endpoint mappings or
manual eligibility rules.

The manual-reference comparison is a distinct evaluation: live agents did not
receive the manual adapters, selection policies, prepared tables, or reference
outputs. Thus it is not a comparison of the generated application to itself.
However, both implementations were developed by the authors on these patient cohorts
and can share interpretation errors. Numerical verification, reference
agreement, and external clinical adjudication are different evidence levels;
only the first two are reported. Independent eligibility validation would
require blinded source-to-output review, including the disputed record selections,
sources with held endpoints, endpoint definitions, and time origins.

The archived offline comparison requires identical, unique monthly grids
and compares risk, event, censor, and survival values. DFS and CRC-specific
survival agree at all evaluated points. OS differs by at most
0.0011329628355009058 in survival probability. The experiment summary
reports nine record-selection differences: four stage-0 GSE39582 cases and four
prior-treated TCGA records held only by the agent, and one TCGA\_COAD case
included only by the agent because of primary-case interpretation.
The net difference is seven fewer patients and two fewer events in the agent
population. These are policy-sensitive selection disagreements, not nine
independently established agent errors. Similar aggregate curves can coexist
with different inclusion policies, so adjudicating these boundaries is a
separate task from verifying the product-limit calculation.

The summary reports matching duration/event targets within numerical
tolerance for records included by both methods. Patient rows, crosswalks,
and local adapters are absent from the compact archive, so this
record-level claim and its discrepancy attribution rely on reported evidence.
Remaining differences include covariate coverage, disposition precedence,
and representation conventions; Section~\ref{sec:tabular-reproduction}
specifies the replay boundary.

\paragraph{Histogram release scope.}
The evaluated application sends per-site event/censor histograms in plaintext
without a minimum-cell or tail-risk-set suppression rule. All three archived
aggregate curves contain singleton event and censor bins and reach risk sets
of one; the two single-client endpoints receive no cross-site pooling
protection. Small patient-cohort size is not itself the disclosure test.
GSE35896's endpoint is held, so none of its 62 source records enter analysis. Bin width is
an analysis setting, not a privacy guarantee. Private deployment would require
a consortium-defined release policy covering sparse cells, tails, and repeated
queries; coarsening, suppression, secure aggregation, or calibrated noise
would need separate evaluation. No such protection or reidentification test
is claimed here.

\subsection{Reproduction and next evidence}
\label{sec:tabular-reproduction}

From the branch's experiment directory,
\texttt{python -B results/compare\_curves.py} reproduces the aggregate
comparison without patient inputs, paid agent calls, or a new simulation.
It was rerun during manuscript preparation with the values in
the main paper's survival-results table. Full preparation needs the exact source
selector, pinned runtime/bridge versions, and omitted site-local artifacts.
The compact archive does not include the complete execution environment or
all site-local artifacts; aggregate replay is not full workflow reproduction.

The summary describes earlier information-pool and policy revisions.
Only the final screened fresh run and manual V5 curves supply the results
reported here. In particular, older documents that include endpoints at GSE35896 or
GSE37892, or hold all GSE14333 records, must not supply final counts.
The execution status retains revision-worklist fields after completion;
completion is corroborated by the accepted ten-site review and all three
persisted analysis results, rather than inferred from that worklist.

Stronger evidence requires repeated held-out runs, independent clinical
adjudication of population boundaries, source-to-output record comparisons,
documentation/contract/review ablations, and matched human preparation with
ordinary coding assistance. Record human preparation and communication
person-hours separately from machine cost and elapsed data-readiness
time. Correctness must be matched before claiming efficiency; no observed
human-time savings are established by this experiment.

\section{Imaging Training Curves}
\begin{figure*}[tp]
\centering
\includegraphics[width=0.94\textwidth]{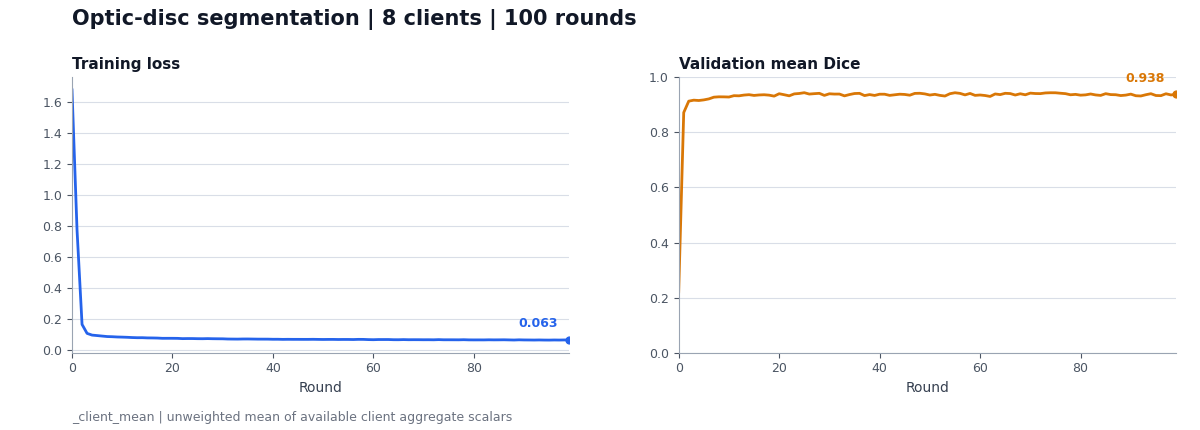}\\
\includegraphics[width=0.94\textwidth]{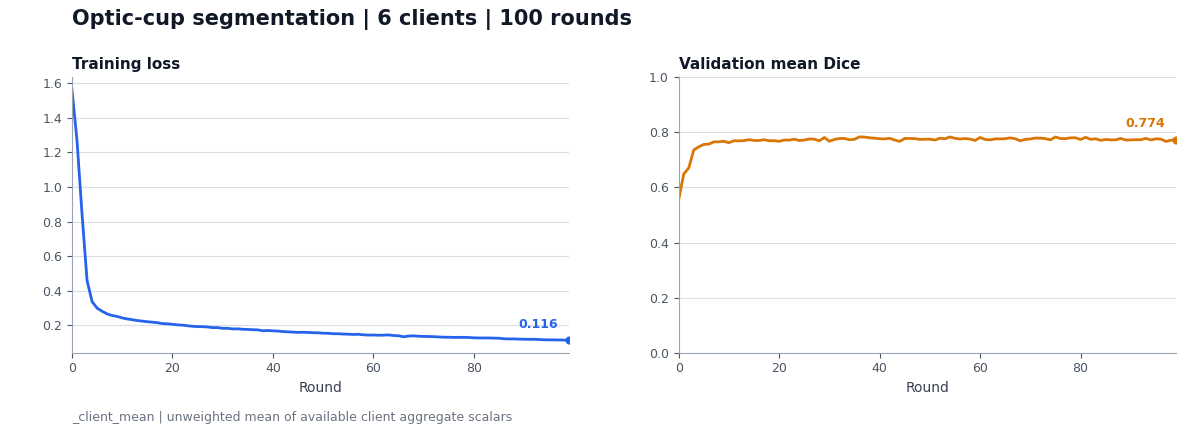}
\caption{Historical segmentation curves: disc (top) and cup (bottom).
Each run, with fixed participating sites, completes
100 rounds of sample-weighted FedAvg with no agent in the loop. Values are
unweighted means of available client scalars for both loss and validation,
not pooled patient-level performance or the sample-weighted training
objective. The curves document downstream execution rather than a controlled
comparison of prediction quality.}
\label{fig:curves}
\end{figure*}

Best validation values were disc Dice 0.9444 at round 25 and cup Dice 0.7839
at round 36.

\begin{figure*}[tp]
\centering
\includegraphics[width=0.94\textwidth]{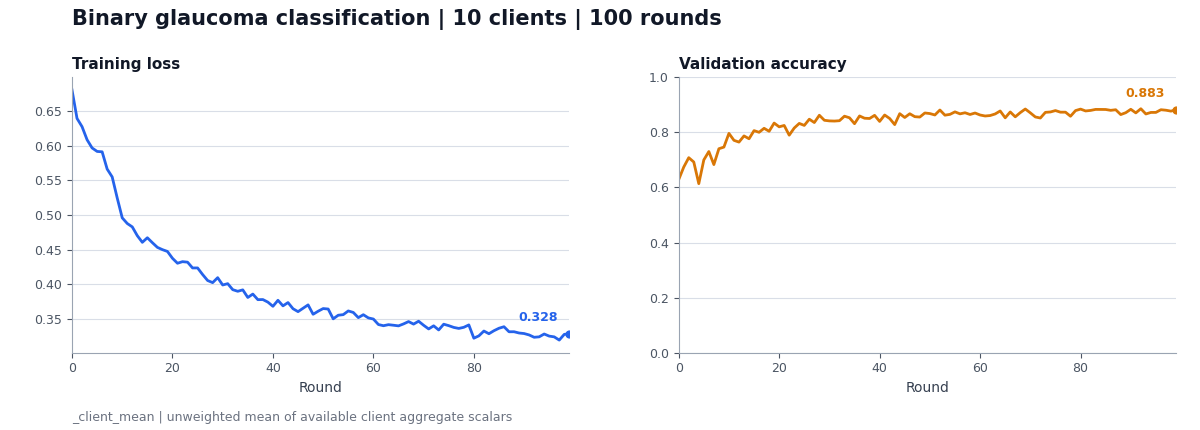}\\
\includegraphics[width=0.94\textwidth]{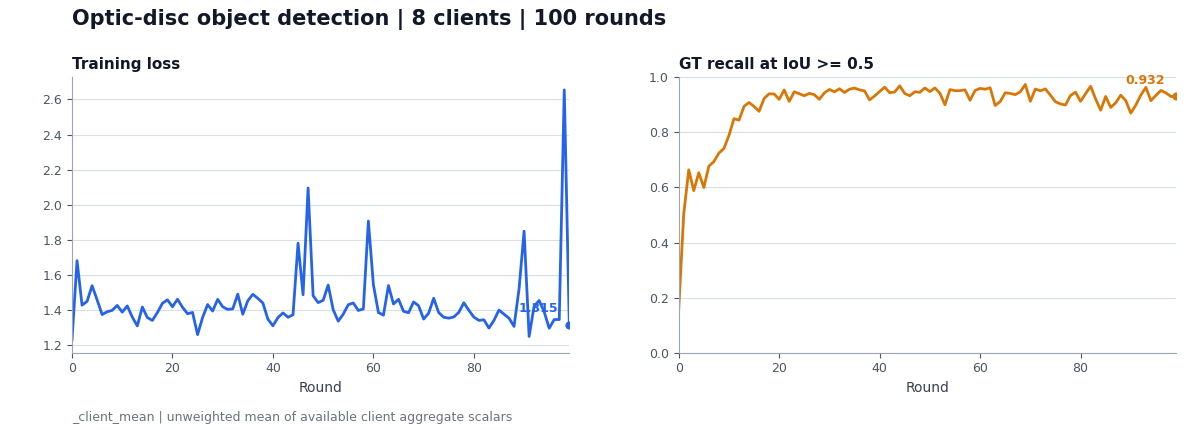}
\caption{Historical classification (top) and detection (bottom) curves,
using the same client-mean convention as Figure~\ref{fig:curves}.
Detection's round-96 communication warning did not prevent later rounds or
persisted terminal artifacts. Temporal proximity does not establish that
the warning caused the loss excursions; no such attribution is claimed.}
\label{fig:classification-detection-curves}
\end{figure*}

Best validation values were classification accuracy 0.8861 at round 92 and
detection GT recall@0.5 0.9742 at round 69; final classification balanced
accuracy was 0.8372. A matched predictive comparison would hold participating sites,
splits, trainer, initialization, and training budget fixed while changing
preparation, and report patient-weighted and client-mean scores with
per-round contributing-client counts.

\end{document}